\documentclass[preprints,article,accept,pdftex,moreauthors]{Definitions/mdpi} 
\firstpage{1} 
\pubvolume{1}
\issuenum{1}
\articlenumber{0}
\pubyear{2026}
\copyrightyear{2026}
\datereceived{ } 
\daterevised{ } 
\dateaccepted{ } 
\datepublished{ } 
\usepackage[utf8]{inputenc}
\usepackage{textcomp}
\usepackage{amsmath}
\usepackage{newunicodechar}
\usepackage{xcolor}
\newunicodechar{∞}{$\infty$}
\newunicodechar{ϵ}{\epsilon}
\DeclareUnicodeCharacter{2212}{\textminus}

\Title{Diffusion-Based Feature Denoising and Using NNMF
for Robust Brain Tumor Classification}

\Author{Hiba Adil Al-kharsan $^{1}$\orcidA{}, Róbert Rajkó $^{2,3}$*\orcidB{}}

\AuthorNames{Hiba Adil Al-kharsan, and Róbert Rajkó }

\address{%
$^{1}$ \quad Doctoral School of Computer Science, University of Szeged, \'{A}rp\'{a}d t\'{e}r 2, H-6720 Szeged, Hungary; hibaadil@inf.u-szeged.hu\\
$^{2}$ \quad University Research and Innovation Center (EKIK), \'{O}buda University, B\'{e}csi \'{u}t 96/b, H-1034 Budapest, Hungary; rajko.robert@uni-obuda.hu \\
$^{3}$ \quad Academic Staff, Doctoral School of Computer Science, University of Szeged, \'{A}rp\'{a}d t\'{e}r 2, H-6720 Szeged, Hungary; rajko@sol.cc.u-szeged.hu
}

\corres{Correspondence: rajko.robert@uni-obuda.hu; EKIK OE}

\abstract{
Brain tumor classification from magnetic resonance imaging, which is also known as MRI, plays a sensitive role in computer-assisted diagnosis systems. In recent years, deep learning models have achieved high classification accuracy. However, their sensitivity to adversarial perturbations has become an important reliability concern in medical applications. This study suggests a robust brain tumor classification framework that combines Non-Negative Matrix Factorization (NNMF or NMF), lightweight convolutional neural networks (CNNs), and diffusion-based feature purification. Initially, MRI images are preprocessed and converted into a non-negative data matrix, from which compact and interpretable NNMF feature representations are extracted. Statistical metrics, including AUC, Cohen’s d, and p-values, are used to rank and choose the most discriminative components. Then, a lightweight CNN classifier
is trained directly on the selected feature groups.
To improve adversarial robustness, a diffusion-based feature-space purification module is introduced. A forward noise method followed by a learned denoiser network is used before classification. System performance is estimated using both clean accuracy and robust accuracy under powerful adversarial attacks created by AutoAttack. The experimental results show that the proposed framework achieves competitive classification performance while significantly
enhancing robustness against adversarial perturbations.The findings presuppose that combining interpretable NNMF-based representations with a lightweight deep approach and diffusion-based defense technique supplies an effective and reliable solution for medical image classification under adversarial conditions.}

\keyword{Brain Tumor Classification; Non-Negative Matrix Factorization (NNMF); Diffusion-Based Defense; AutoAttack+ (AA+); Metrics for Classification Models; CNN-based Classification}

\begin{document}

\maketitle

\section{Introduction}
The classification of brain tumors from magnetic resonance imaging (MRI) is a large and complex component in computer-supported diagnostic systems.
Early and careful detection improves handling, design, and patient survival. In recent years, deep
learning approaches, mostly convolutional neural
networks (CNNs), have shown remarkable performance in the medical
image analysis job~\cite{litjens2017survey}. CNN-based models
have high classification accuracy in brain tumor detection and segmentation problems due to their ability to
learn hierarchical features directly from data.

Despite their powerful predictive performance, deep neural
networks are highly vulnerable to adversarial attacks.
Little, carefully crafted input modifications can safely
degrade classification accuracy while remaining visually
imperceptible~\cite{szegedy2013intriguing}. This sensitivity 
raised serious concerns in safety-critical applications, such as medical diagnosis, where solidity and robustness are essential. To ensure a robust evaluation of the robustness of the framework,
united attack benchmarks such as AutoAttack have been suggested~\cite{croce2020autoattack}.

Feature extraction~\cite{ref_feature_selection} and dimensionality reduction remain key components for building interpretable and computationally efficient classification models. Similar chemometric approaches such as PLS-DA have been successfully applied in other domains, illustrating the value of interpretable low-dimensional representations.~\cite{RAJKO2024e35045}.
Non-negative matrix factorization (NNMF or NMF) is a well-confirmed
method to decompose non-negative data into representations based on collective parts~\cite{lee1999nmf}. 
Unlike unconstrained  linear techniques such as Principal
Component Analysis (PCA), NNMF step non-negativity
bands, resulting in an interpretable low-rank string.
This property makes NNMF especially suitable for medical
image analysis, where the pixel density and derived features
are inherently non-negative~\cite{ref_nmf_book}. 

In this work, we suggest a structured framework for
brain tumor classification that combines NNMF-based feature extraction, statistical feature ranking, CNN-based classification, and diffusion-based feature purification for adversarial robustness.
Firstly, MRI images are converted into non-negative feature matrices and decomposed using NNMF.
The extracted components are statistically evaluated using metrics such as Area Under the Curve (AUC), Cohen's $d$, and hypothesis testing to identify the most spatial features.
Then, a lightweight CNN classifier is trained on the selected feature subset.
To improve robustness, a feature-space diffusion is introduced to affect structured noise injection, followed by a learned denoising
network that borders the reverse diffusion operation. System performance is evaluated against strong adversarial attacks using AutoAttack~\cite{croce2020autoattack}, with the implementation adopted from the official public store~\cite{autoattack_repo}. The estimate compares baseline and defended models in terms of both
clean precision and robust precision.
The suggested path demonstrates that the
combination of interpretable NNMF representations with lightweight
deep models and diffusion-based purification provides competitive classification accuracy while improving robustness versus
adversarial attacks. {
However, current methods, such as PCA, autoencoders, and transformer-based embeddings, likely lack interpretability in medical applications. In contrast, NNMF provides a parts-based representation that is proper to its non-negativity constraints, making it more suitable for modeling tumor structures in brain MRI images.
Moreover, while many studies focus on improving classification accuracy, little attention has been paid to robustness to adversarial perturbations. This work aims to address this hole by merging diffusion-based feature denoising with interpretable feature extraction.
}

\section{Related Works}
\begin{itemize}
{

\item Deep Learning-Based Brain Tumor Classification (Recent Advances) \\ 
Almuhaimeed et al.~\cite{ref_almuhaimeed_2025} proposed a deep learning framework for brain tumor classification using MRI images, integrating convolutional neural networks with data augmentation techniques. Their model achieved high classification accuracy exceeding 97\%, demonstrating the effectiveness of deep learning approaches in medical imaging.
However, their evaluation was conducted under standard conditions without considering adversarial robustness.
In contrast, the current study evaluates both clean and adversarial performance using AutoAttack.

\item Hybrid CNN and Transformer Models \\ 
Gómez et al.~\cite{ref_gomez_2025} proposed a hybrid CNN–Transformer architecture for multi-class brain tumor classification. Their model combines local feature extraction with global attention mechanisms, achieving improved classification performance.
However, this approach increases model complexity and reduces interpretability.
In contrast, the proposed framework employs NNMF to generate interpretable feature representations.

\item Diffusion and Generative Models in Medical Imaging \\ 
Deem et al.~\cite{ref_deem_2026} analyzed robustness in brain tumor classification using modern deep learning models and highlighted the trade-off between classification accuracy and adversarial robustness.
Similarly, recent studies have explored generative approaches such as GANs and diffusion models to improve performance and robustness.
However, most of these methods operate in pixel or latent space.

}
 \item {Brain Tumor Detection Using CNN}
\\Hossain et al.~\cite{ref_cnn_brain} suggest a hybrid brain tumor 
detection framework that joins classical machine learning classifiers 
with a computational neural network (CNN). The research used the 
BRATS benchmark dataset and applied Fuzzy C-Means (FCM) clustering 
for tumor segmentation, followed by feature extraction using texture 
and statistical descriptors.

In the classical machine learning phases, several classifiers were 
evaluated, including Support Vector Machine (SVM), K-Nearest Neighbors 
(KNN), Logistic Regression, Naïve Bayes, Random Forest and 
Multilayer Perception (MLP). between them, SVM achieved the best 
performance with an accuracy of 92.42\%.

For deep learning-based classification, the researcher designed a 
five-layer CNN structure depending on convolution, max-pooling, 
flattening, and fully linked layers. The suggested CNN achieved 
97.87\% accuracy in an 80:20 training–testing split, 
outperforming traditional classifiers.

However, the performance reported increases; the approach relies on 
pixel-level segmentation and explicit image-based CNN classification. 
In contrast, our work achieves interpretable low-rank NNMF 
representations combined with a lightweight CNN approach and 
diffusion-based robustness boost, aiming not only at high 
classification accuracy but also improved adversarial robustness.
    \item NMF-CNN for the enhancement of
 features \\Chan et al.~\cite{ref_nmf_cnn_dcase} suggest a hybrid NMR-convolutional neural network (NMF–CNN) framework for the detection of sound events in the DCASE 2019 challenge. In their work, NMF was used as a preprocessing and feature enhancement step to approximate powerful labels from weakly labeled data by resolving the analysis matrix $H$. The extracted representations were then fed into a CNN approach for event classification.
The results of their study showed that the integration of NMF with a shallow 
CNN enhances the event-based F1-score (30.39\%) compared to the baseline 
system (23.7\%), explaining that NMF can provide a meaningful structural decay that benefits deep learning architectures.
Unlike their system in voiced scene analysis, the current study 
adopts NNMF for medical image feature extraction, where it is applied to generate interpretable low-rank representations of brain MRI images. 
Subsequently, these representations are utilized for classification and robustness evaluation under autoattacks

\item Semi-NMF Network for Image Classification

Huang et al.~\cite{huang2019seminmf} suggest a Semi-NMF-based
convolutional network (SNnet) for image classification, where
Convolutional filters are not learned through backpropagation 
but instead are built using semi-non-negative matrix factorization
used to image patches.
Unlike traditional CNNs that depend on slope-based optimization,
Their process learns filter banks by matrix factorization,
reduces computational value, and avoids global parameter tuning.
 In addition, a weakly supervised extension (S-SNnet) was introduced
via merge graph regularization into the Semi-NMF framework
to enhance discriminative capability.
Experimental results on the MNIST dataset demonstrated that
The suggested style achieves competitive performance compared to
state-of-the-art shallow and deep learning architectures such as PCANet.
This work highlights the feasibility of merging the matrix
factorization mechanism within convolutional frameworks for
active feature learning.
    \item {Classification-Denoising Joint Models}

Thiry and Guth~\cite{thiry2024classification} suggest classification-denoising
 networks, which simultaneously model image classification and
denoising by learning a single network that holds the common distribution of noisy image information and their labels. Their method
combines loss of cross-entropy classification with the proper denoising outcome
 at multiple noise levels, using the Tweedie–Miyasawa
formula to estimate the denoised product. Experimental results in
CIFAR-10 and ImageNet demonstrate competitive performance and
improved adversarial robustness compared to standard discriminative
classifiers. This study provides
a theoretical link between
denoising objectives and adversarial gradients, offering a new view of robustness that complements conventional defense
mechanisms~\cite{thiry2024classification}.

    \item {Reliable Robustness Evaluation}

Croce and Hein~\cite{croce2020autoattack} propose a robust
evaluation framework for adversarial robustness based on a crew of various parameter-free attacks. Unlike earlier
benchmarks that often build on individual attack procedures,
Their ensemble joins complementary attacks such as APGD
and Square attacks to provide a parameter-independent and
safe robustness rating. The evaluation of the study was 
tested on more than fifty models and explained that many
 Already considered robust defenses could be broken when
value with the suggested attack ensemble. This study
highlighted the need for strict and united robustness
estimate in adversarial machine learning and directly
Motivated the use of AutoAttack as a reliable benchmark in the current work.
\item {
Recent studies (2024–2026) on brain tumor classification have reported high precision using the deep CNN model and transformer-based models, particularly in publicly available datasets. Many of these approaches focus mainly on maximizing clean accuracy, often exceeding 95\%. More three reviews~\cite{AHAMED2023102313, SystRevHybridML2025, app16020831} treated brain tumor segmentation, potential of hybrid models and analysis of model interpretability, resp.

However, these methods are typically evaluated under classical conditions and do not consider adversarial robustness or interpretability. In contrast, the suggest work confirms robustness against adversarial perturbations while preserving interpretable NNMF-based feature representations. }
\end{itemize}
Thus, this study provides a complementary perspective by addressing reliability and robustness, which are critical in safety-sensitive medical systems.

{
In addition to classification-based approaches, recent studies have discussed anomaly detection and the mechanism of medical image segmentation for brain tumor testing, with the aim of improving localization accuracy and robustness.

The suggested work differs by focusing on interpretable feature extraction combined with adversarial robustness over diffusion-based feature purification, providing a complementary perspective to existing methods.
}

\section{Materials and Methods}
\begin{figure}
    \centering
    \includegraphics[width=0.8\textwidth]{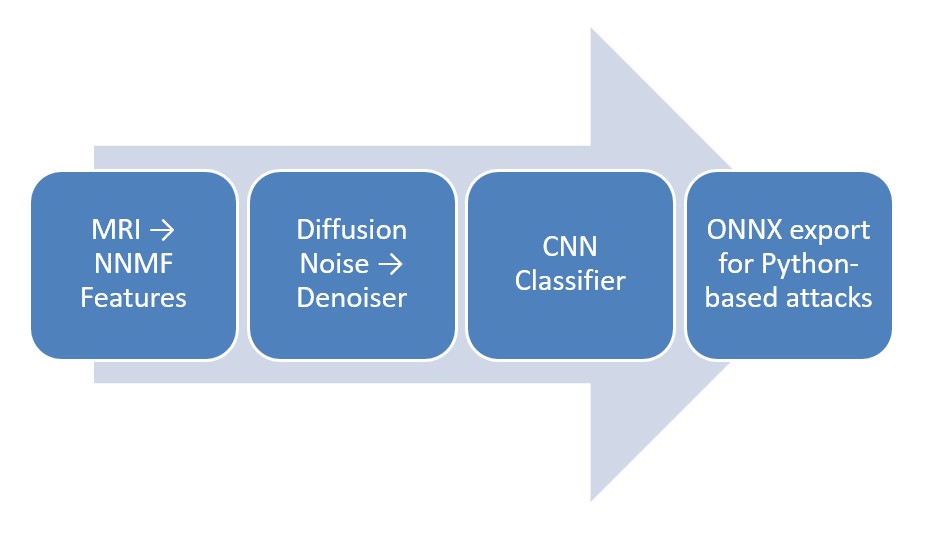}
    \caption{Stages of the proposed framework}
    \label{fig:stages}
\end{figure}

 To develop a robust and adversarial classifier, a structured sequence of fundamental phases is required for the proper implementation of a machine learning model. In this work, a neural network–based model is adopted as the core classification algorithm. The proposed classifier is constructed through four main stages, each comprising a set of well-defined sub-processes prepared to ensure computational correctness, stability, and robustness against adversarial attack.
 {
The functions of NNMF and CNN in the suggested pipeline are complementary rather than redundant. At first tempt, CNN provided excellent classification, however AutoAttack destroyed it totally, because of the full overfitting, that is why NNMF was applied to extract low-rank and interpretable feature representations from MRI data, hold meaningful structural patterns.

CNN runs on the selected NNMF features as a lightweight classifier, focusing on discriminative learning rather than feature extraction from raw images. This separation minimizes model complexity and avoids redundancy between components and overfitting.

The diffusion module is used in the feature space after NNMF and feature selection, acting as a purification step that enhances robustness against adversarial perturbations. It operates independently of the feature extraction stage and does not interfere with the interpretability of the NNMF representations. Therefore, each component in the pipeline serves a distinct and non-overlapping role.
} These four stages are shown in Figure~\ref{fig:stages}, which {
 summarizes the overall workflow of the suggested framework, including preprocessing, feature extraction based on NNMF, CNN-based classification, diffusion-based denoising and robustness evaluation.
}
\subsection{Preprocessing data set }
The data set was selected from Kaggle, a well-known platform for data science and machine learning research.
This data set contains brain magnetic resonance images with identical segmentation masks, which are usually used to train and evaluate brain tumor segmentation models~\cite{dataset}. This data set consists of approximately 2,200 brain magnetic resonance images, each with a binary segmentation mask that highlights the tumor part at the pixel level. It is intended for semantic segmentation function and has been mostly used in research and public notebooks on Kaggle to train and evaluate deep learning models such as U-Net and its variants for brain tumor analysis. It was originally provided in the "COCO annotation"  format, where the images and their corresponding labels were stored separately in a JSON file. The first preprocessing script in Python was developed to reorganize the data set into a folder directory structure suitable for image classification tasks. The script parses the annotation file, maps each image to its corresponding category, and saves the images to dedicated folders that represent each class: 730 training, 219 validating and 97 testing items for normal images, and 770 training, 210 validating and 118 testing items for tumor images. Thus, the train/validation / test ratios were kept and the single original Kaggle split was used only~\cite{dataset}, i.e., 70\% / 20\%/ 10,5\%], respectively. There was no exact information on the number of slices per patient, neither for the patient-wise representation. Because patient identification is not available, slice-level leakage between splits cannot be fully ruled out, and reported performance may be optimistic compared to a strictly patient-wise evaluation. See details on the risk of cut-level leakage in brain magnetic resonance imaging classification in~\cite{ref_leakage_paper}. {
As we know, there is no published classification examination of this Kaggle brain tumor dataset.} 

These preprocessing steps enable for smooth integration with convolutional neural network training.

{
It is important to note that the data set used in this study is derived from a semantic segmentation task and does not include patient identifiers. Therefore, data separation was performed at the slice level rather than at the patient level.

This may introduce a possible risk of data leakage, leading to an optimistic performance rating. However, all preprocessing and splitting procedures were applied consistently in all experiments to give fair evaluation.
Future work will consider patient-wise data splitting and additional sensitivity analysis to further validate the robustness and generalization of the proposed framework.
}

\subsection{Extraction Feature Phase Using NNMF}
In this phase, features are extracted from each image using
Non-Negative Matrix Factorization (NNMF). NNMF is a mechanism to factorize a non-negative matrix 
$V \in \mathbb{R}^{K \times N}_{+}$ into two non-negative matrices
$W \in \mathbb{R}^{K \times R}_{+}$ and
$H \in \mathbb{R}^{R \times N}_{+}$, so that their product
approximates $V$, as reported in~\cite{ref01}.

\begin{equation}
V = WH + E
\end{equation}
where $E \in \mathbb{R}^{K \times N}$ is the reconstruction error matrix.
The matrices $W$ and $H$ are estimated by minimizing the cost function
between $V$ and $WH$ as follows~\cite{ref01}:

\begin{equation}
W = \arg\min_{W} C(V \mid WH), \quad \text{for fixed } H
\end{equation}

\begin{equation}
H = \arg\min_{H} C(V \mid WH), \quad \text{for fixed } W
\end{equation}
where $C(A \mid B)$ indicates a distance measurement between the matrices $A$ and $B$.
Various distance measures can be used, such as the Euclidean distance,
the Kullback--Leibler divergence, the Itakura--Saito divergence, and
$\beta$-divergence.

NNMF is used not only as a dimensionality-lowering mechanism,
but also as an interpretable low-rank approximation model for
non-negative data. As clarified in the SIAM study on
Non-negative Matrix Factorization~\cite{ref_nmf_book}, NMF belongs
to the family of linear dimensionality lowering mechanism,
where every sample appears as an additive group of
a small number of basis components under non-negativity chain's.
Unlike PCA or unconstrained low-rank approximations, the
non-negativity constraint executes parts-based representations,
which are particularly proper for image analysis and medical data
where pixel density is inherently non-negative.

Moreover, the option of an objective function significantly affects
the obtained decomposition and its statistical interpretation
~\cite{ref_nmf_book}. Consequently, in the suggested framework,
NNMF is optimized using the Kullback--Leibler (KL) divergence with
multiplicative update rules~\cite{ref02}:

\begin{equation}
W \leftarrow W \otimes
\frac{\left( V ./ (WH) \right) H^{T}}
{\mathbf{1}_{K \times N} H^{T}}
\tag{4}
\end{equation}

\begin{equation}
H \leftarrow H \otimes
\frac{W^{T} \left( V ./ (WH) \right)}
{W^{T} \mathbf{1}_{K \times N}}
\tag{5}
\end{equation}
where $\otimes$ and $./$ denote element-wise multiplication and division,
respectively, and $\mathbf{1}_{K \times N}$ represents a matrix $K \times N$ 
whose elements are all equal to one.\par
\begin{figure}
    \centering
    \includegraphics[width=0.85\textwidth]{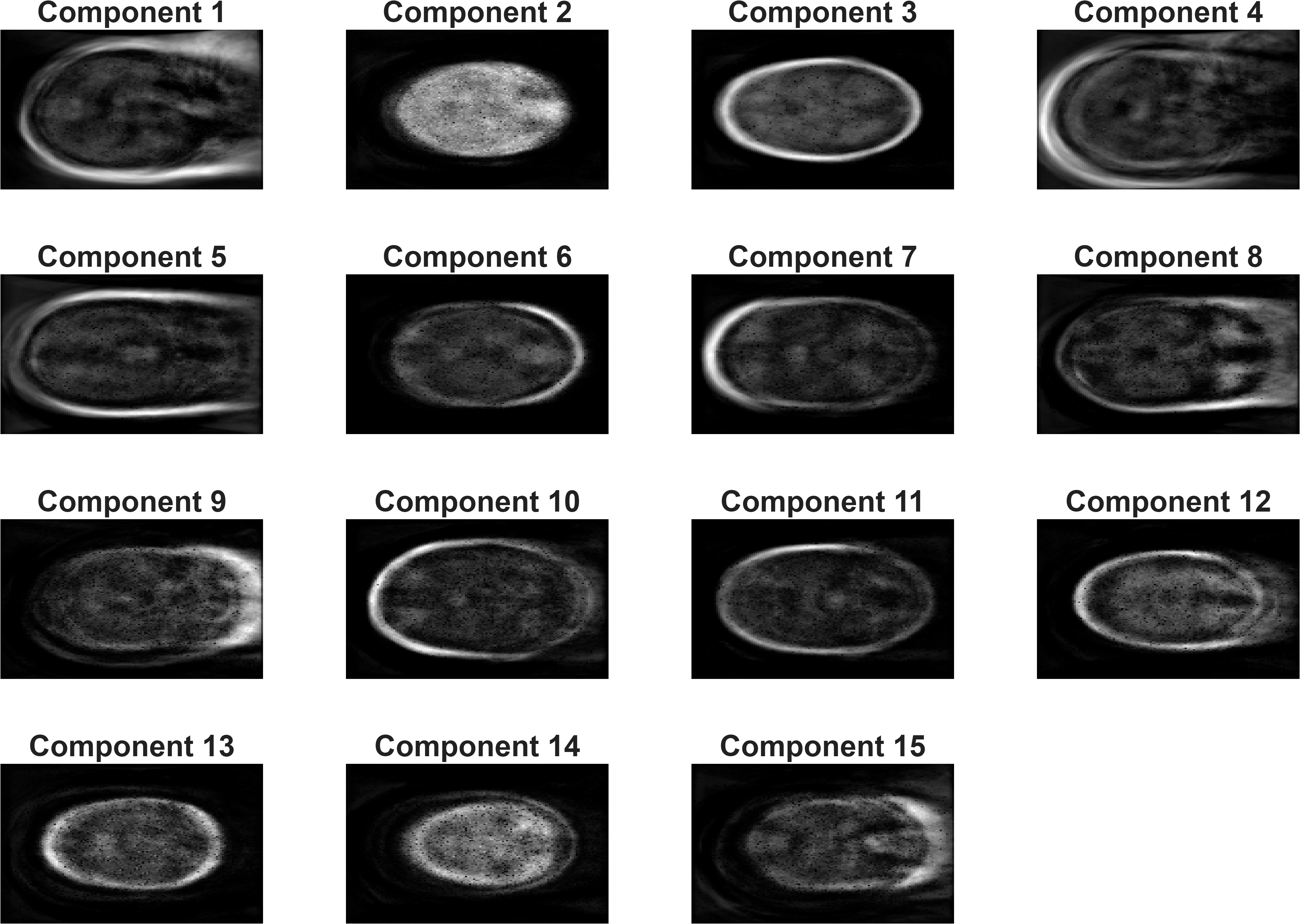}
    \caption{NNMF Basis Components (k = 15) .
This figure shows the learned NNMF basis components got from the training data. Each basis image represents a non-negative spatial pattern that contributes to rebuilding brain MRI images. The components take meaningful anatomical structures such as skull boundaries, tissue distribution, and localized density variations. The variety across components indicates that NNMF decomposes the images into multiple complementary patterns rather than a single dominant structure, providing an interpretable and compact representation suitable for later analysis.}
    \label{fig:nnmf_feature1}
\end{figure}

\begin{figure}
    \centering
    \includegraphics[width=0.85\textwidth]{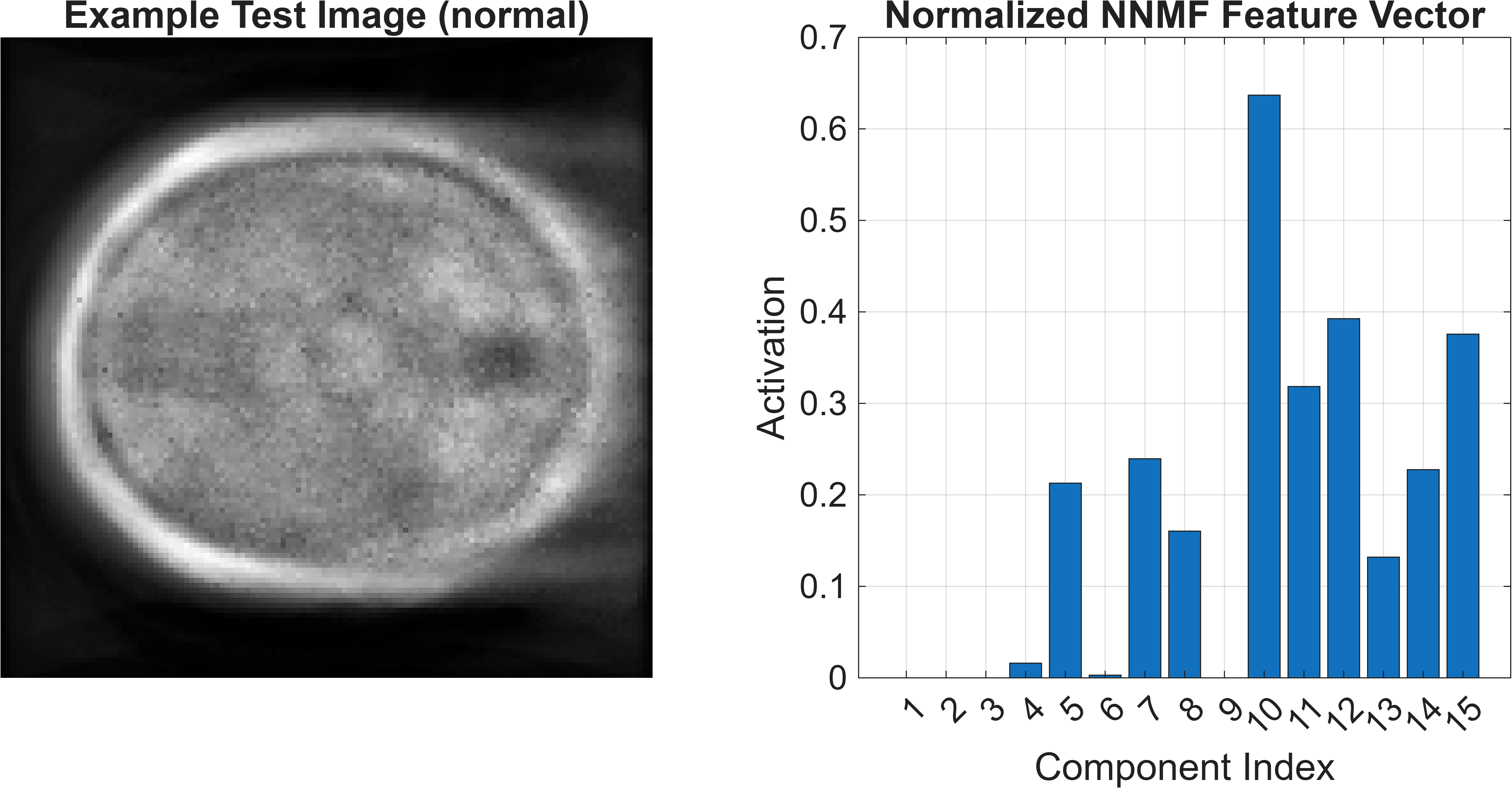}
    \caption{Example Test Image and Its Normalized NNMF Feature Vector.
    This figure illustrates an example test MRI image from the normal class side, its matching normalized NNMF feature vector. The bar plot shows the activation power of each NNMF component for this image, highlighting that only a subset of components exhibits strong responses. This sparse and selective activation pattern explains that NNMF features encode discriminative structural information rather than uniformly responding to all components, which is desirable for robust classification.}
    \label{fig:nnmf_feature2}
\end{figure}

\begin{figure}
    \centering
    \includegraphics[width=0.85\textwidth]{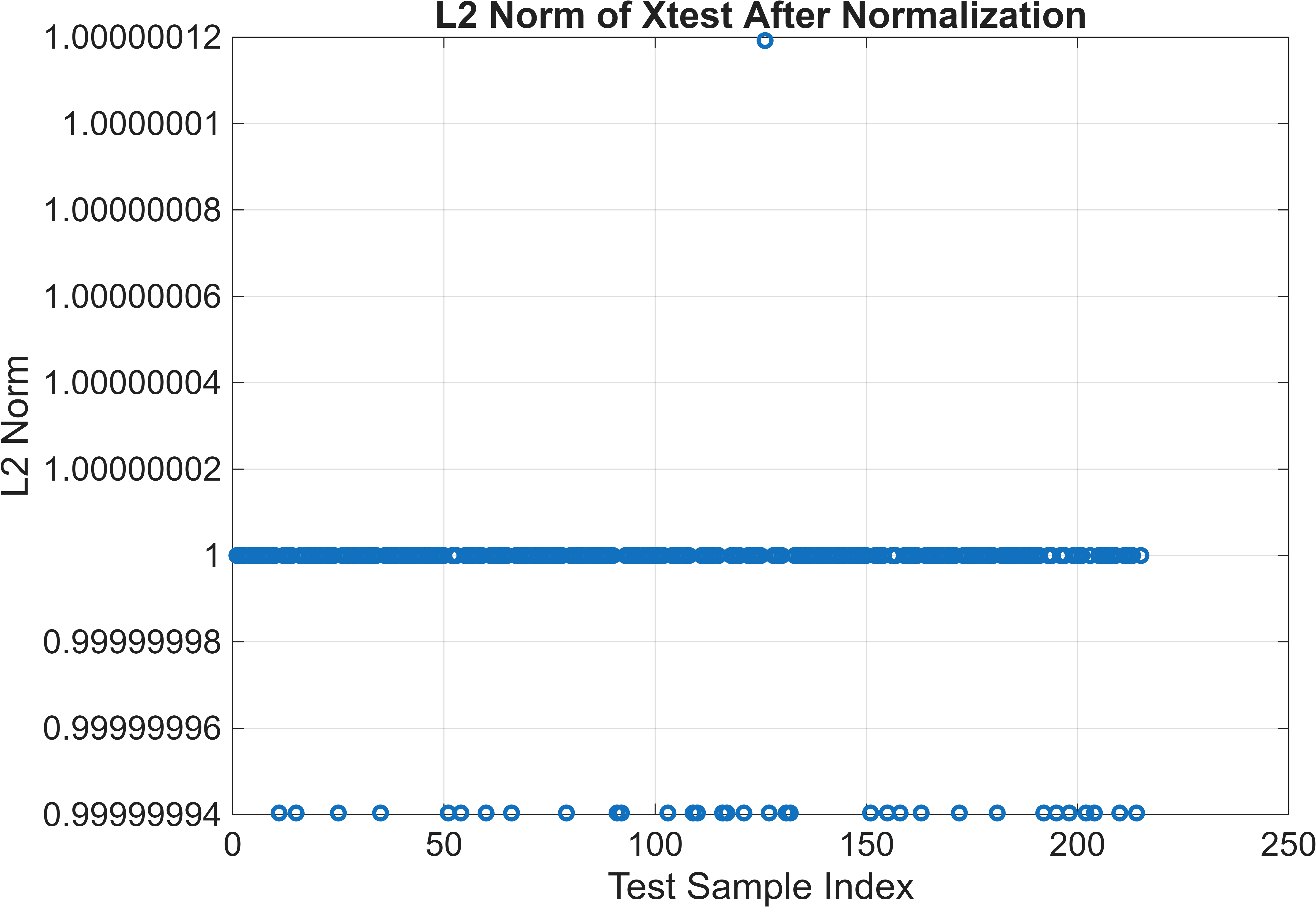}
    \caption{L2 Norm of Xtest After Normalization.
This figure reports the L2 norm of all normalized test feature vectors. The values are tightly focused around one, confirming the right and stability of the normalization process. Ensuring unit-norm feature vectors is critical for fair comparison between samples and for robustness evaluation, as it blocks feature magnitude variations from controlling the classifier or adversarial attack.}
    \label{fig:nnmf_feature3}
\end{figure}

\begin{figure}
    \centering
    \includegraphics[width=0.85\textwidth]{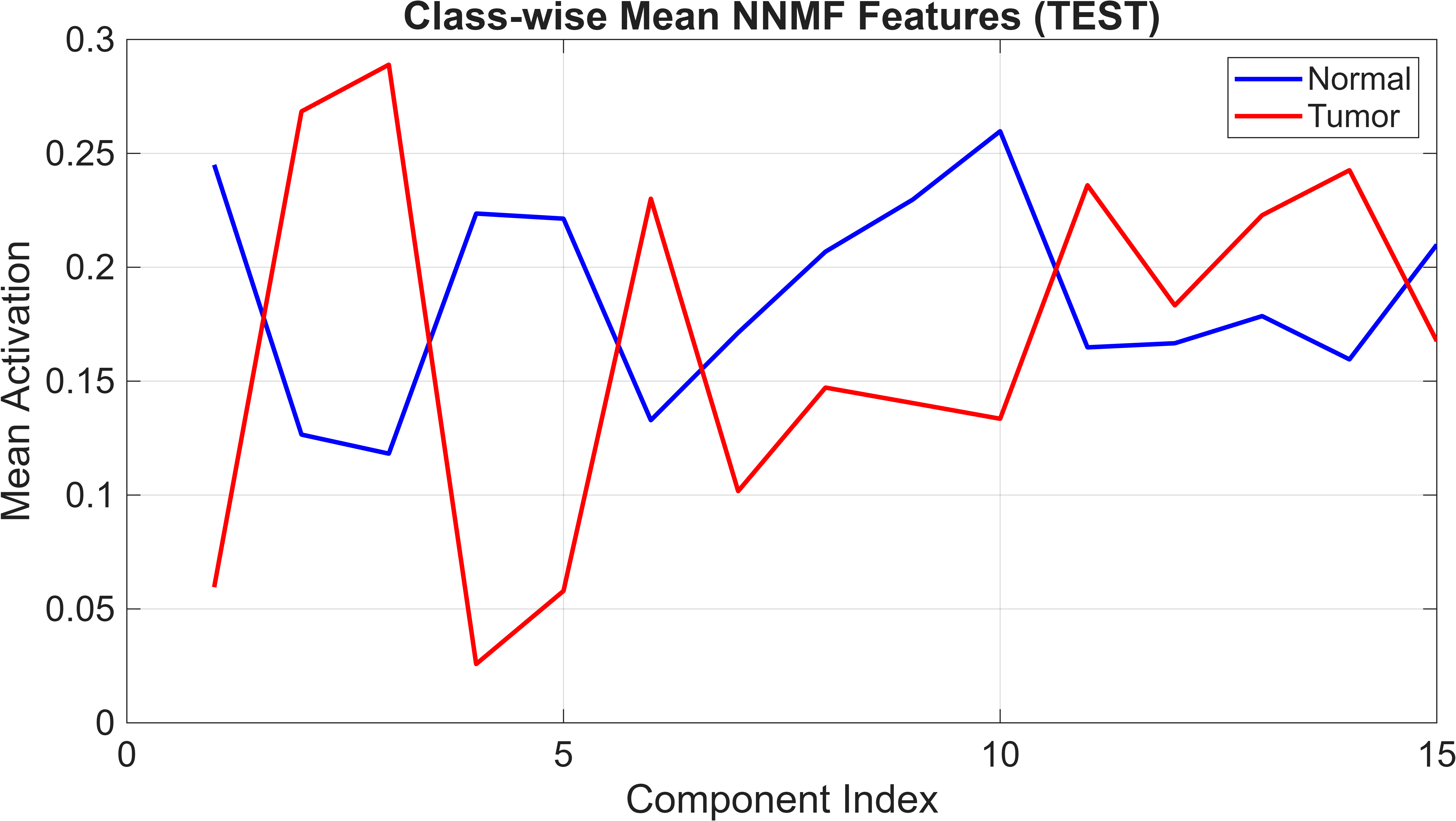}
    \caption{Class-wise Mean NNMF Features (Normalized, TEST).
This figure shows the mean activation of each NNMF component for normal and tumor classes after feature normalization. Clearly, differences can be noticed via several components, where certain features show continuously, higher activation for tumor samples, while others are more prominent for normal samples. These class-dependent activation patterns indicate that NNMF components capture discriminative characteristics linked to pathological changes, supporting their effectiveness for classification and feature-space defense strategies.}
    \label{fig:nnmf_feature4}
\end{figure}

\begin{figure}
    \centering
    \includegraphics[width=0.85\textwidth]{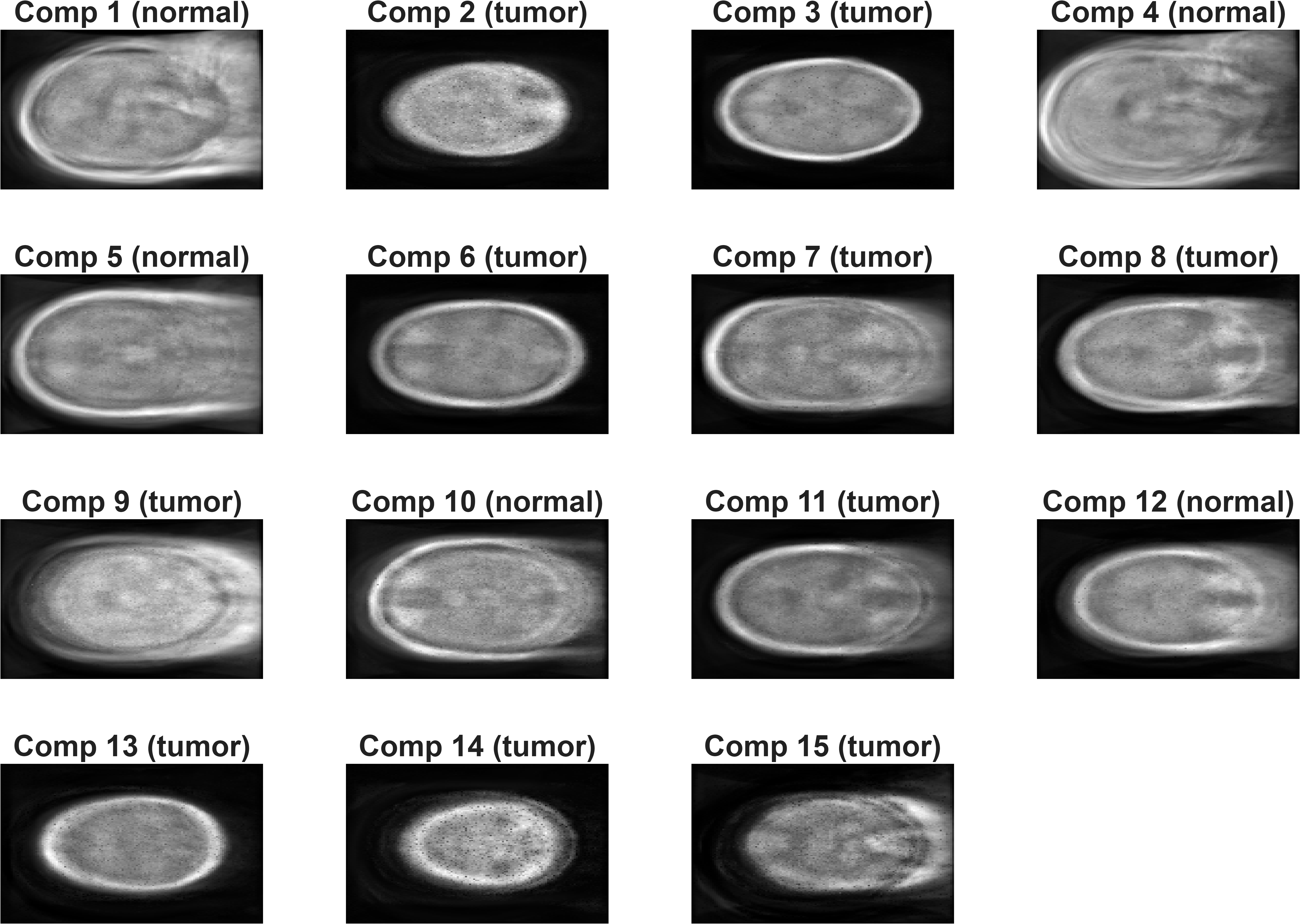}
    \caption{Top-Activated TEST Samples per Component (Using Normalized Xtest:
This figure appears to be the most strongly activated test sample for each NNMF component based on normalized feature vectors. For each component, the corresponding image and its activation value are viewed along with the class label. The results detect that some components are often activated by tumour images, while others respond more strongly to normal brain structures. This component-wise association boosts interpretability by directly linking devoid feature dimensions to concrete visual patterns in MRI images
}.
    \label{fig:nnmf_feature5}
\end{figure}

 The images were loaded using MATLAB Datastore with labels class from folder names. The image was transformed to grayscale if necessary, resized to 128×128, normalized to the [0,1] range, and vectorized to form a non-negative data matrix V, where each column represents one image. NNMF was learned using the training set, producing a basis matrix W and a coefficient matrix H. Validation and test features were got by drop their vectors onto the fixed basis W via non-negative least squares. Finally, the feature vectors were L2-normalized and saved for classification and robustness. Figures ~\ref{fig:nnmf_feature1}-~\ref{fig:nnmf_feature5} explain the steps to apply this method:
{
Figure~\ref{fig:nnmf_feature1} presents the learned NNMF basis components and shows that the decomposition captures multiple meaningful structural patterns. Figure~\ref{fig:nnmf_feature2} gives an example of a normalized NNMF feature vector for a test image, while Figure~\ref{fig:nnmf_feature3} confirms the stabilization of the L2-normalization process. In addition, Figures~\ref{fig:nnmf_feature4} and~\ref{fig:nnmf_feature5} illustrate class-wise activation behavior and the most powerfully activated test samples, respectively, supporting the interpretability of the extracted NNMF features.
}


\subsection{Feature Selection and Statistical Feature Analysis}
\begin{figure}
    \centering
    \includegraphics[width=0.85\textwidth]{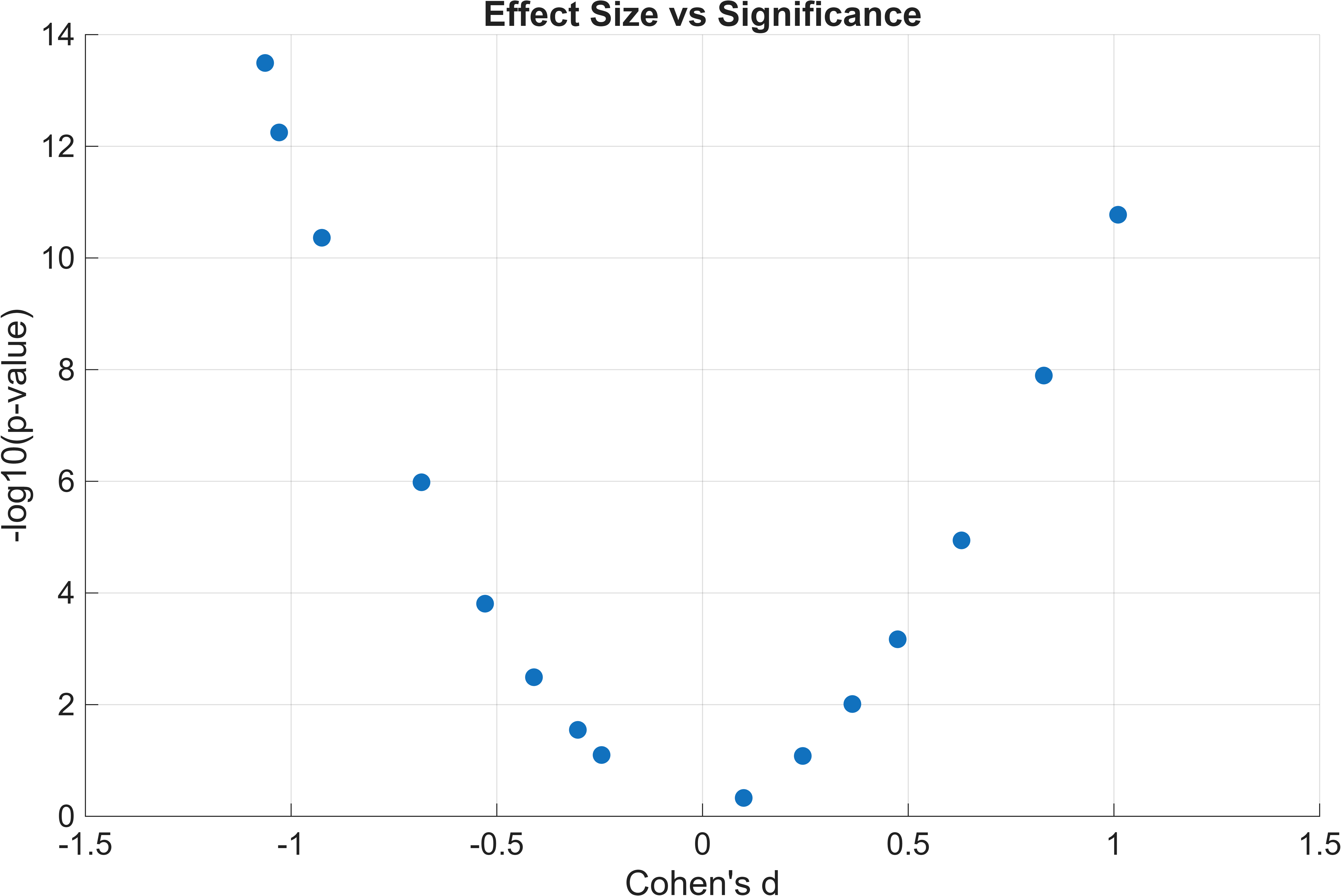}
    \caption{Effect Size vs Significance.
This figure clarifies the relation between effect size and statistical significance for NNMF features. The horizontal axis marks Cohen’s d, where positive values correspond to higher activation in tumor samples and negative values point to higher activation in normal samples. The vertical axis performs the negative logarithm of the p-value gained from Welch’s t-test. Features with large absolute effect sizes and high statistical significance are visually stressed, highlighting components that are both hardly discriminative and statistically trusted.}
    \label{fig:nnmf_feature6}
\end{figure}

Statistical analysis was performed on the NNMF features extracted from brain magnetic resonance images to identify the most distinguishing components to classify tumors and normal tissues, the NNMF rank was identified as 15. Each feature of the NNMF was individually evaluated using multiple complementary criteria, including the area under the ROC curve (AUC), the size of the effect measured by Cohen's d-value and the statistical significance assessed using Welch’s t-test. This multi-criteria evaluation allows for an accurate result for the features by jointly considering discriminatory power, class separation size, and statistical reliability. The selected features provide interpretability and a concise representation suitable for subsequent classification and robustness analysis. {
Figure~\ref{fig:nnmf_feature} highlights the discriminative power of the selected components of the NNMF based on AUC values, while Figure~\ref{fig:nnmf_feature6} shows the relationship between impact size and statistical significance. The allocation plots and heatmap further support the hypothesis that the selected features capture meaningful class-dependent differences between normal and tumor samples.
}
\begin{figure}
    \centering
    \includegraphics[width=0.85\textwidth]{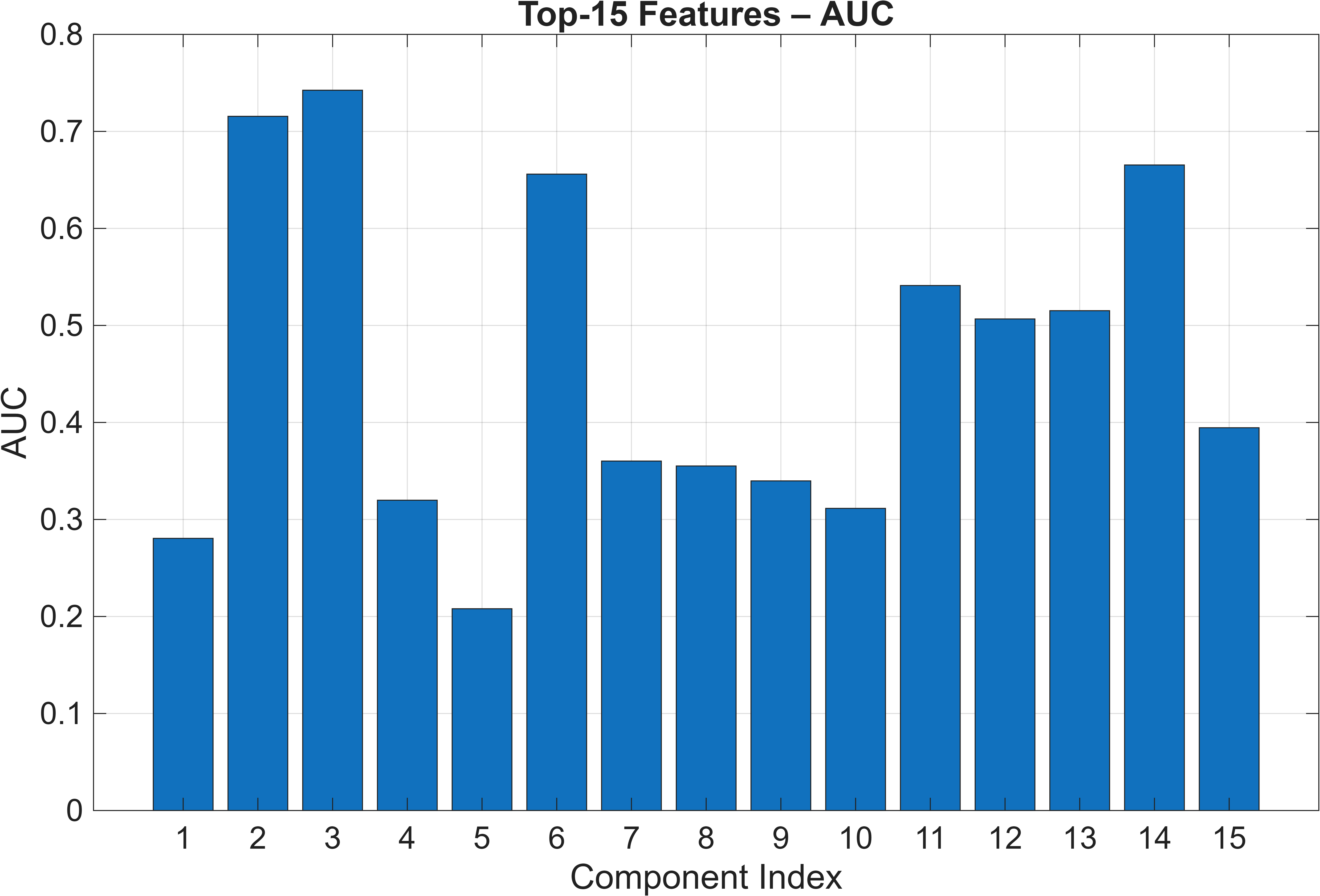}
    \caption{Top-15 Features – AUC.
This figure shows the AUC rate of the top rate of NNMF features based on the feature selection operation. Each bar corresponds to a single NNMF component and affects its individual ability to distinguish tumor samples from normal ones. Higher AUC numbers indicate the most powerful discriminative ability, while values closer to 0.5 suggest limited separability. The score proves that several NNMF components show meaningful classification potential at the feature grade.}
    \label{fig:nnmf_feature}
\end{figure}
\begin{figure}
    \centering
    \includegraphics[width=0.85\textwidth]{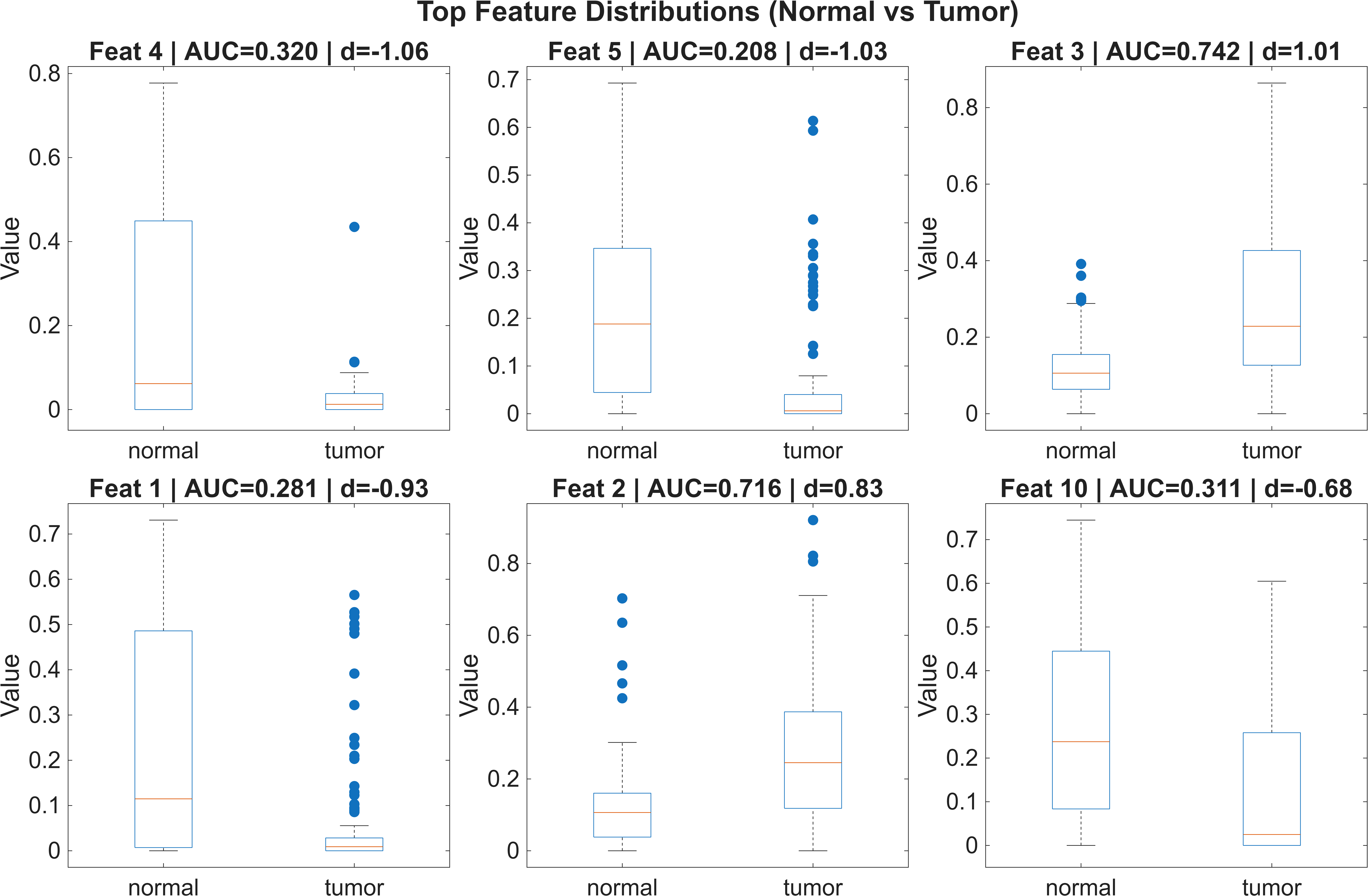}
    \caption{Top Feature Distributions (Normal vs Tumor)
This figure displays boxplot visualizations of the maximum discriminative NNMF features, comparing their normalized distributions between normal and tumor classes. The plots detect visible differences in average and allocation spreads for selected features, providing visual confirmation of their discriminative stand. These distributions supplement the statistical metrics and explain how individual NNMF components react differently to healthy and sick brain structures.}
    \label{fig:nnmf_feature}
\end{figure}
\begin{figure}
    \centering
    \includegraphics[width=0.85\textwidth]{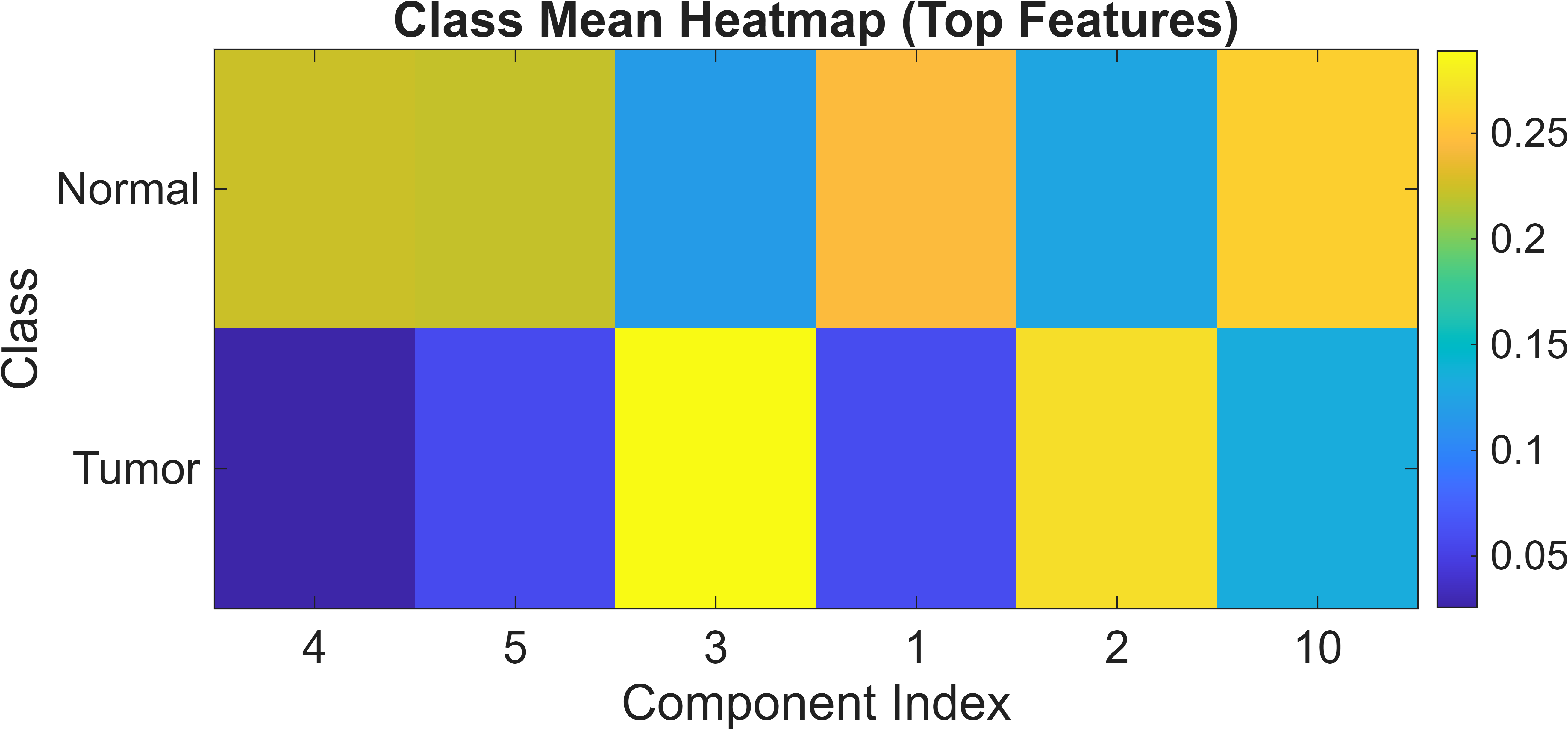}
    \caption{Class Mean Heatmap (Top-15 Features)
This figure summarizes the class-wise mean activation of the top rate NNMF features using a heatmap representation. Every column reacts to a selected NNMF component, while rows represent the normal and tumor classes. The color intensity reflects the average feature activation, enabling speedily identification of tumor-dominant and normal-dominant features. The observed style confirms that NNMF components capture class-based structural information and contribute to interpretable feature-level discrimination.}
    \label{fig:nnmf_feature}
\end{figure}
{
Of the components initially extracted from NNMF (k = 15), the final subset of features used for classification was selected based on a combination of the AUC ranking, Cohen's effect size d, and statistical significance (p-values). The Top-M features (M = 15) were selected and consistently used in training, validation and test sets.
}
\subsection{CNN-Based Classification on Selected NNMF Features}
After extracting features using NNMF, a statistical feature ranking stage is applied to evaluate the discriminative force of each NNMF component. Specifically, features are rated using multiple statistical criteria, including the Area Under the ROC Curve (AUC), Cohen's effect size d, and p-values derived from hypothesis testing. These metrics quantify class separability and enable the selection of the most informative NNMF components while refusing redundant or weakly discriminatory features.
\\
Based on this ranking, a reduced subset of the top M NNMF features is selected and run as input to a lightweight Convolutional Neural Network (CNN) model. This step aims to explore whether the compact and interpretable NNMF representation is sufficient for proper brain tumor classification without the dependence on high-dimensional image inputs. before training, the selected feature vectors are smooth using L2 normalization per sample basis to stabilize the learning operation and ensure comparable feature scales across all samples. The CNN model is trained using the normalized NNMF features of the training set, while performance is monitored on a validation set to block overfitting. Lastly, the trained model ranks on a held-out test set using standard classification metrics, including accuracy and confusion matrices. This evaluation explains the effectiveness of combining the NNMF-based dimensionality decrease with CNN-based classification, achieving reliable tumor-versus-normal discrimination while maintaining a compact and strong feature representation. Similar hybrid NNMF-CNNs strategies have been shown to be effective in related style recognition and signal processing models, where NNMF supplies interpretable features and CNNs capture nonlinear decision boundaries~\cite{ref02,ref_nmf_cnn_dcase}.
{
Figure~\ref{fig:nnmf_feature05} displays CNN training behavior in terms of accuracy and loss, while the validation and test confusion matrices explain that the classifier is able to recognize between normal and tumor samples with reasonably balanced performance.
}
{
Although traditional classifiers such as SVM, Random Forest, and Gradient Boosting are usually used for low-dimensional data, CNN was selected in this work to preserve compatibility with deep learning-based pipelines and to enable seamless integration with the diffusion-based defense technique.
Future work may include benchmarking against classical machine learning models to provide a more comprehensive comparison.}

\begin{figure}
    \centering
    \includegraphics[width=0.85\textwidth]{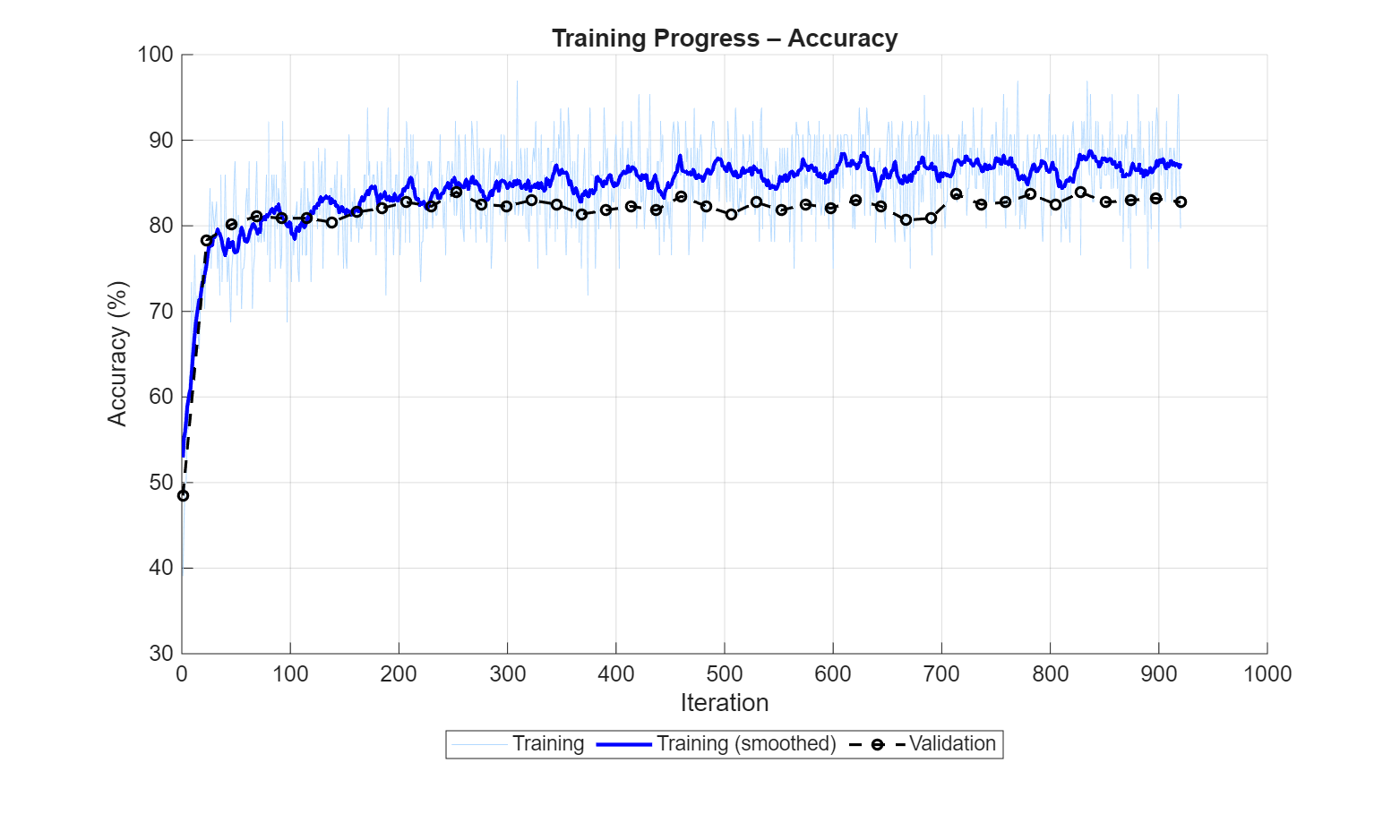}
    \caption{Training Progress – Accuracy
This figure shows the evolution of classification accuracy via training. The light curve represents raw mini-boost training accuracy, while the smoothed curve highlights the total way. Validation accuracy (black markers) is measured periodically and close to the training curve, indicating stable learning and limited overfitting. Accuracy climbs sharply during the early iterations and then gradually saturates, proposing convergence of the model when trained on the selected NNMF feature space.}
    \label{fig:nnmf_feature05}
\end{figure}
\begin{figure}
    \centering
    \includegraphics[width=0.85\textwidth]{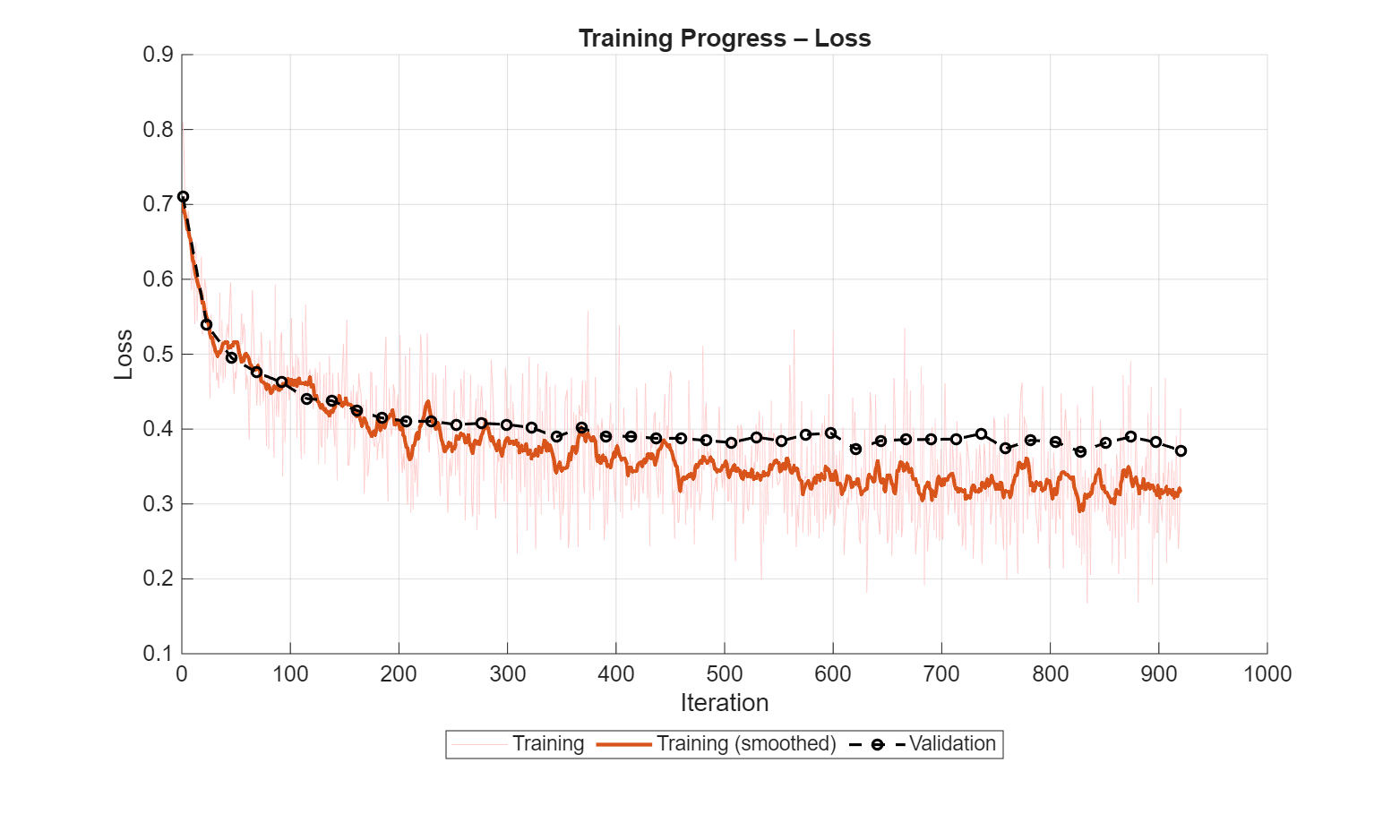}
    \caption{Training Progress – Loss
This figure reports the training and validation loss curves over iterations. The training loss reduces steadily, while the validation loss follows a similar downward trend with a slightly higher amount, which is expected. The parallel behaviour of both curves suggests that optimization is forward normally ,and the model generalizes reasonably well. The lack shows a big difference between training and validation loss, indicating that the CNN is not severely overfitting in spite of being trained for multiple epochs.}
    \label{fig:nnmf_feature05}
\end{figure}
\begin{figure}
    \centering
    \includegraphics[width=0.85\textwidth]{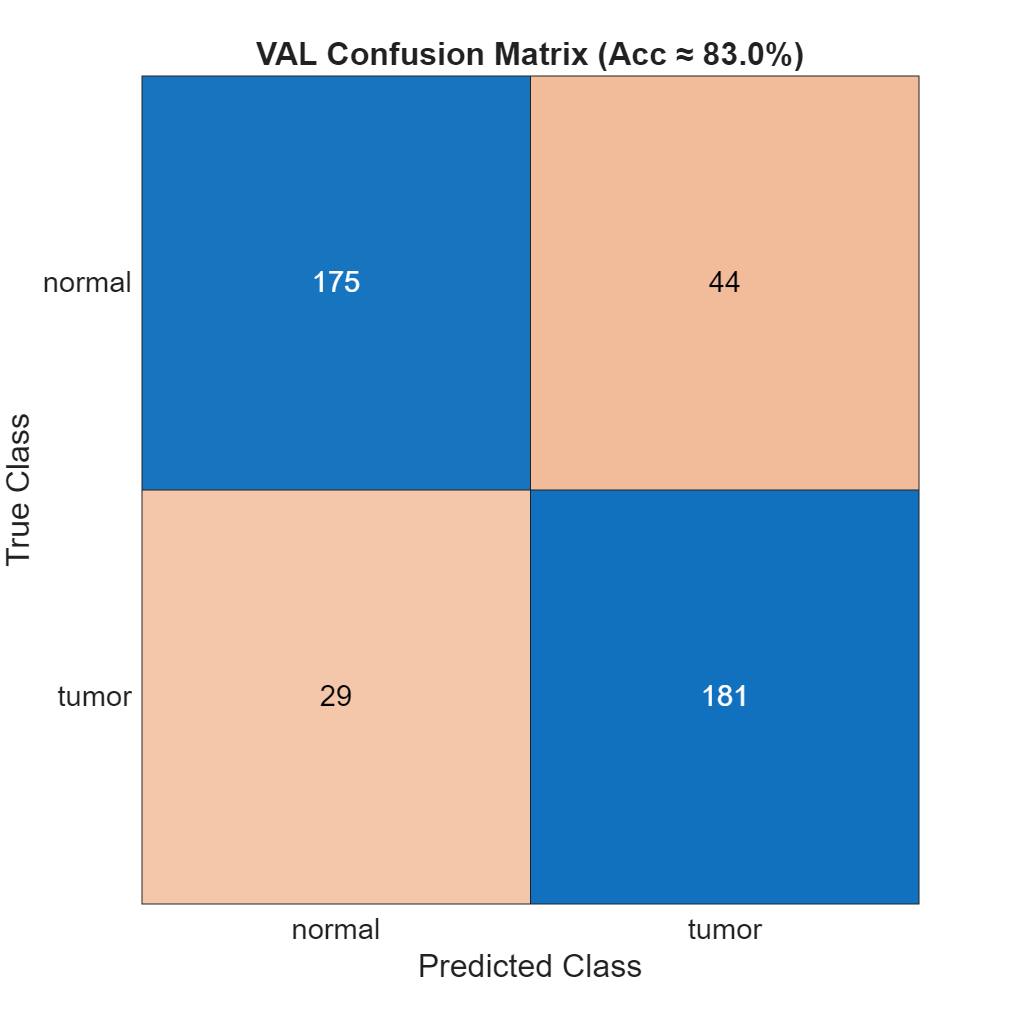}
    \caption{VAL Confusion Matrix (Acc $\approx$ 0.83).
This confusion matrix abstract model's performance on the validation set. Correct predictions show on the diagonal, while off-diagonal input corresponds to m wrong classifications. The matrix points a high rate of correctly classified normal and tumor samples, with errors basis occurring when normal images are predicted as tumor and vice versa. The total validation accuracy (~83\% confirms that the selected NNMF features retain meaningful discriminative information and that the classifier is learning a separable decision limit.}
    \label{fig:nnmf_feature05}
\end{figure}
\begin{figure}
    \centering
    \includegraphics[width=0.85\textwidth]{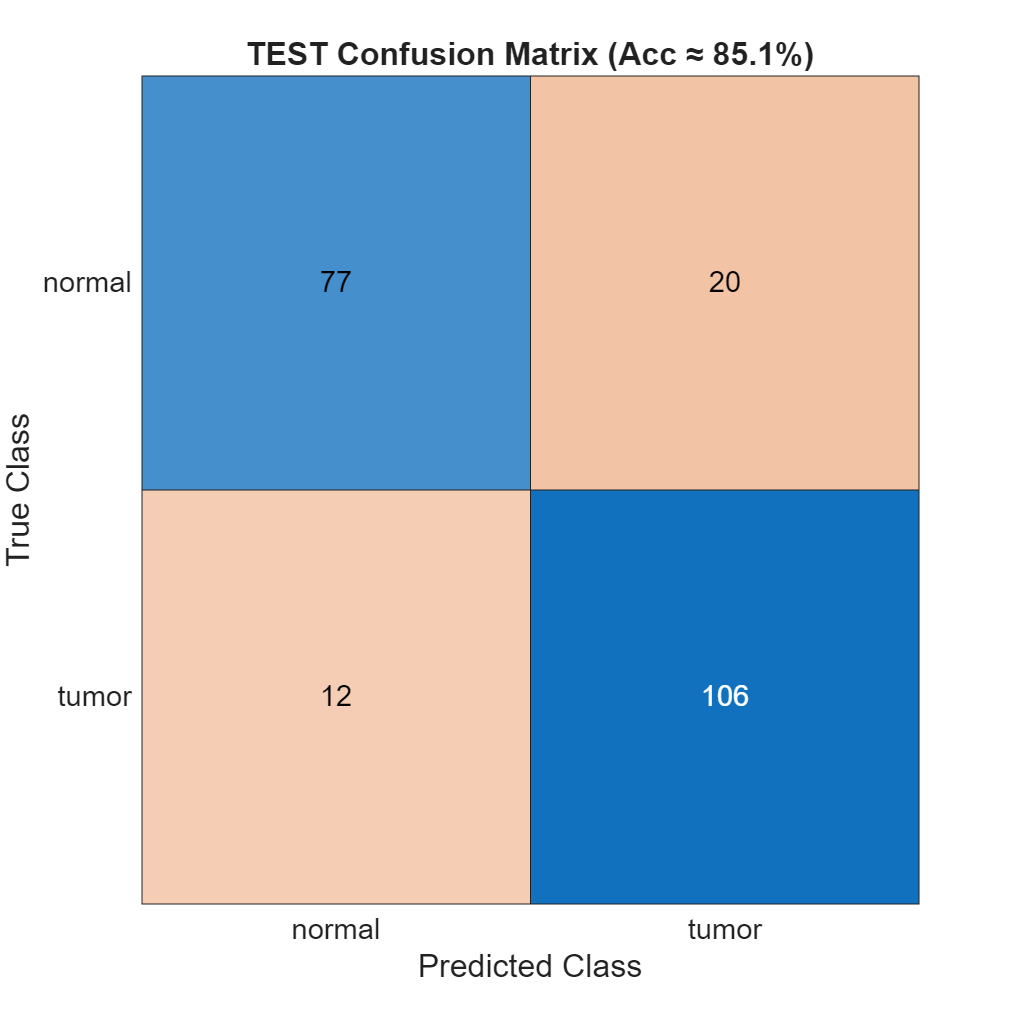}
    \caption{TEST Confusion Matrix (Acc $\approx$ 0.851).
This confusion matrix reports final performance on the unseen test set. The majority of samples lie on the diagonal, return an overall accuracy $\approx$ 0.851. {
Although this accuracy may seem lower compared to some current approaches, it is substantial to confirm that many reported results are gained under standard (non-adversarial) situation and often depend on complex deep architectures. In contrast, the suggested framework prefer robustness and interpretability in supplement to classification performance. The model run on a compact set of statistically chosen NNMF features and uses a lightweight CNN, which improves generalization and minimize overfitting.
Furthermore, under robust adversarial attacks (e.g., AutoAttack), the baseline model display a big drop in performance, while the defended model preserve substantially higher robustness. This demonstrates the effectiveness of the suggested diffusion-based feature purification and highlights the importance of robustness in safety critical implementation such as medical diagnosis.
} The relatively low number of false positives (normal → tumor) and false negatives (tumor → normal) points to balanced performance across both classes. This result supports the effectiveness of using a compact subset of NNMF features (selected by statistical ranking) as input to a small CNN, achieving robust generalization while reducing dimensionality. }
    \label{fig:nnmf_feature05}
\end{figure}

\subsection{Feature-Space Diffusion Data Generation}
In this step, diffusion training data was built in the NNMF feature space to allow feature-level denoising. First, the most discriminative Top-M NNMF features were selected based on feature ratings obtained during the labeling phase. 
All feature ranking statistics were computed exclusively on the training set; the selected Top‑M components were then applied unchanged to the validation and test sets.
To ensure consistency across the pipeline, the same per-sample L2 normalization used by the classifier was applied to all selected features.
A forward diffusion process was then defined using a linear noise timetable with a fixed number of diffusion steps. For each training sample, Gaussian noise was gradually added to the clean feature vector according to the accruing diffusion factor, generating noisy feature representations at randomly selected timesteps. This process yielded paired samples comprising clean features and noisy versions.
The generated diffusion pairs were stored as (xt,x0), where X0 indicates the original clean feature vector, and Xt represents its noisy counterpart in diffusion step t=41. These pairs form the training sets for the feature denoiser in the later step. In addition, the diffusion constants needed for inference and refinement were saved to ensure consistency between the MATLAB and Python implementations.
This step does not involve model training or visualization; instead, it focuses only on setting up structured diffusion data that features representations' phase corruption enables effective feature-space denoising and robust classification in subsequent stages.
The figures below show the attitude of the NNMF features under the forward diffusion process, indicating the phased corruption of the feature representations as the diffusion time step increases.
{
Unlike classical diffusion models, which are mostly used for image generation in pixel or latent space, the suggested approach applies diffusion in feature space as a controlled denoising technique. This project ensures that the operation does not change the semantic structure of the NNMF representations.
Specifically, the diffusion process operates on already extracted interpretable features, introducing noise and then removing it using a learned denoiser. Since the denoising step aims to recover the original feature distribution, the interpretability provided by NNMF is preserved rather than disrupted.
Therefore, the suggested diffusion-based purification boost is robust against adversarial perturbations while maintaining the intrinsic structure and interpretability of the feature space}.
{
The figures in this subsection explain the effect of the forward diffusion process on NNMF features, also the corruption of features at increasing time steps, the change in the distribution of feature-values, and the increase in the noise energy as diffusion proceeds.
}

\begin{figure}
    \centering
    \includegraphics[width=0.85\textwidth]{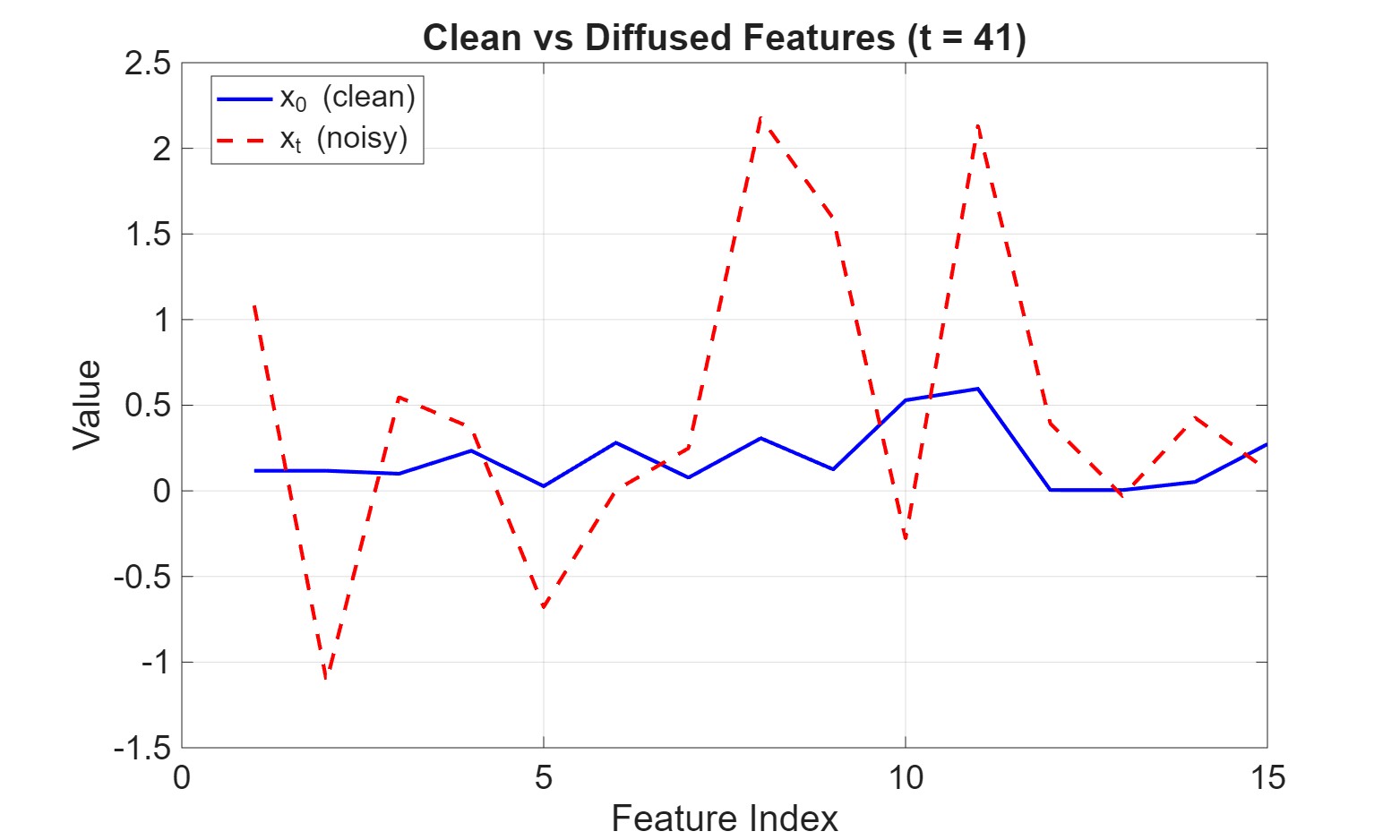}
    \caption{Clean vs diffused NNMF feature vectors at a late diffusion time step (t = 41). The figure shows the impact of forward diffusion noise on selected feature components, view how the original clean features xt,x0  are gradually corrupted to getting noisy features. trusted.}
    \label{fig:nnmf_feature06}
\end{figure}

\begin{figure}
    \centering
    \includegraphics[width=0.85\textwidth]{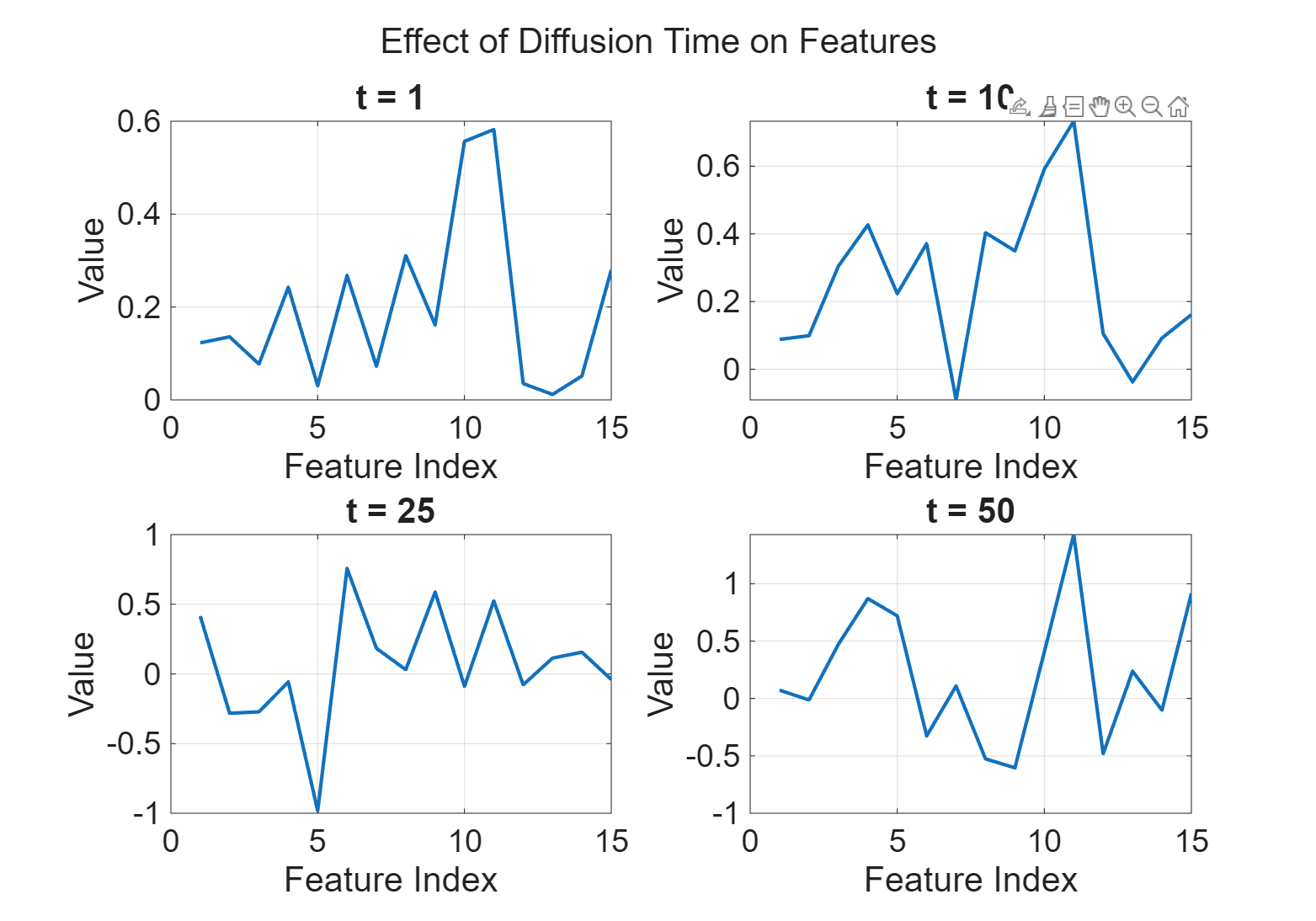}
    \caption{Impact of diffusion time on NNMF features at various timesteps (t = 1, 10, 25, and 50). As the diffusion timestep increases, the injected noise becomes more dominant, leading to higher distortion and variability in the feature representations.}
    \label{fig:nnmf_feature06}
\end{figure}

\begin{figure}
    \centering
    \includegraphics[width=0.85\textwidth]{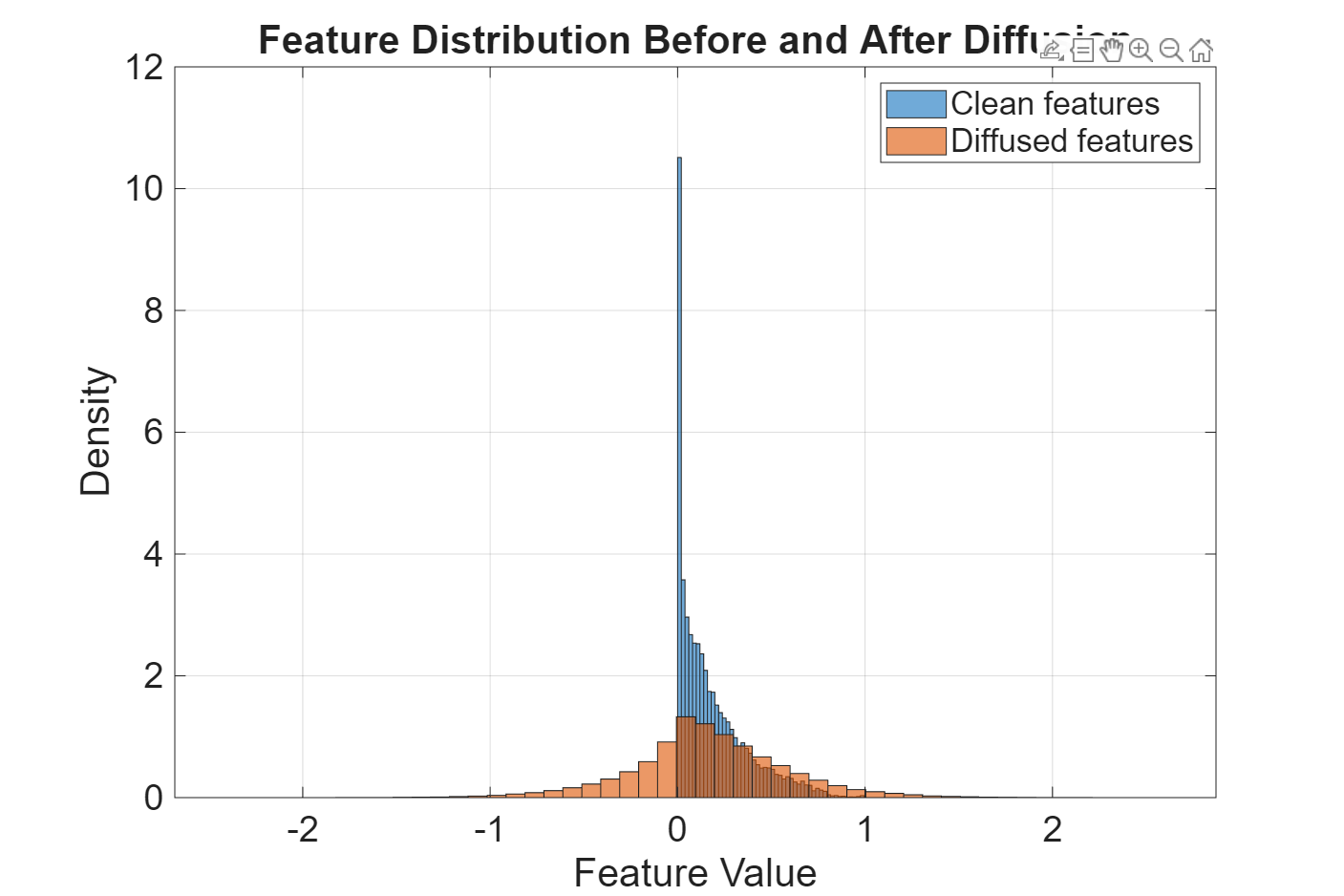}
    \caption{allocation of NNMF feature values before and after diffusion. The histogram comparison highlights the increase in contrast and spread of feature values caused by the diffusion operation, indicating a deviation from the original feature multiple.}
    \label{fig:nnmf_feature06}
\end{figure}

\begin{figure}
    \centering
    \includegraphics[width=0.85\textwidth]{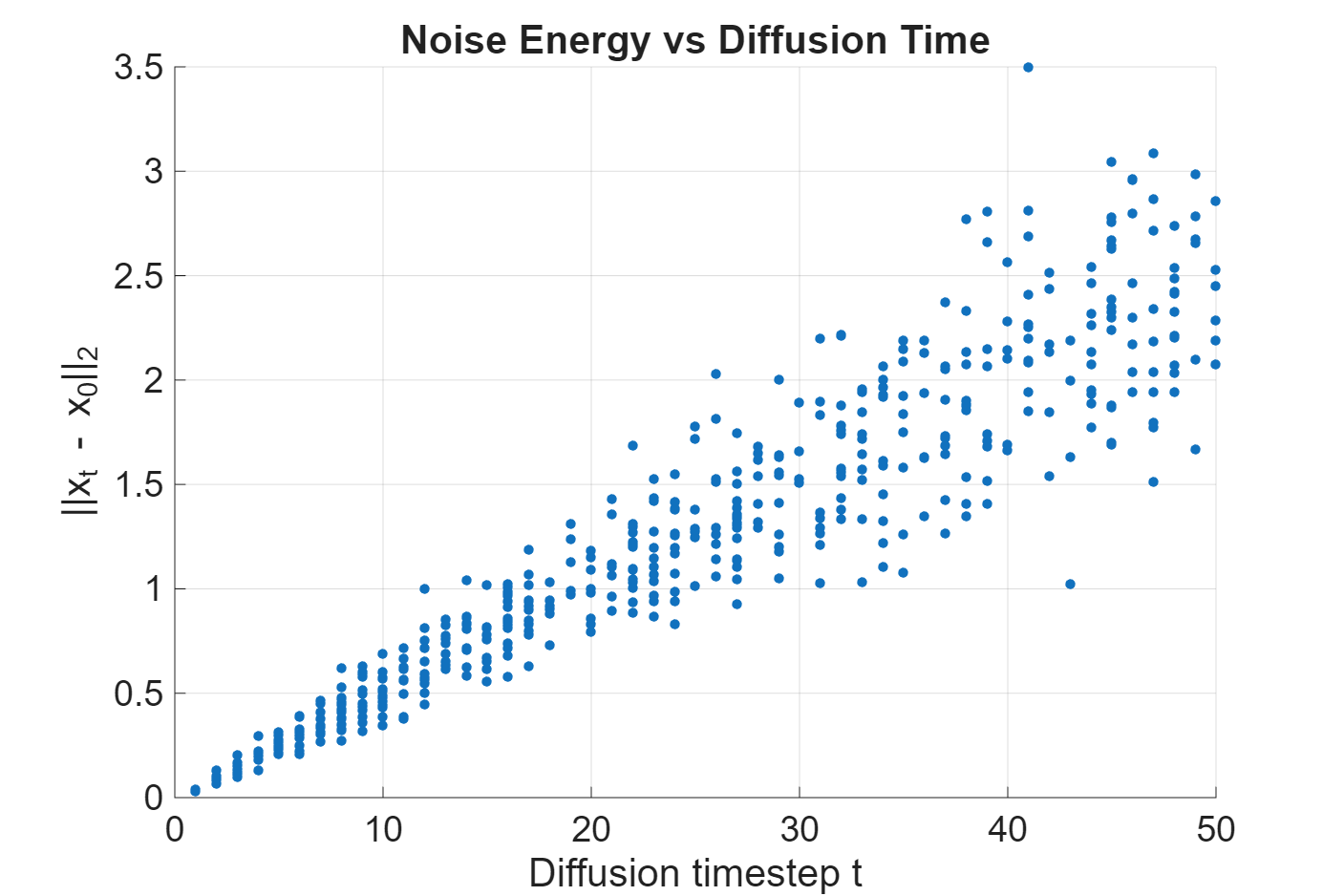}
    \caption{Noise energy as a function of diffusion timestep. The plot shows the L2 distance between clean and noisy feature vectors, $x_t \parallel^2 \parallel x_0$, increasing with diffusion time, quantitative confirming the gradually damage introduced by the forward diffusion step.\\ \newline Although the accumulative diffusion table is not explicitly explained, its effect is inherently reflected in the progressive rise of the noise capacity, which confirms how the contribution of the original clean features gradually reduces as the diffusion timestep boosts.}
    \label{fig:nnmf_feature06}
\end{figure}
{
The diffusion operation is defined using a fixed number of time steps (T = 50) with linear noise. The denoising network is trained to rebuild clean feature vectors from their noisy counterparts.
The timestep information is combined into the denoiser using a normalized scalar embedding, which allows the model to adjust its denoising behavior according to the diffusion step.
The selection of t = 41 for evaluation is based on experimental observations, where this time step provides a balance between suitable noise perturbation and effective recovery by the denoiser.
}

\subsection{Feature-Space Denoiser Training}

After the construction of diffusion-corrupted feature pairs in the previous step, this stage focuses on learning a feature-level denoising model able to recover clean NNMF representations from noisy diffusion samples. The target of this step is to round the reverse diffusion step in the feature space by training a regression-based neural network that maps a noisy feature vector xt, conditioned on its diffusion timestep t, back to its matching clean feature vector x0
To allow for noise-level awareness, the diffusion timestep is coded using a compact sinusoidal embedding and hierarchically with the noisy feature input. The denoiser is optimized using a mean squared error objective, allowing it to gradually suppress diffusion-induced perturbations while keeping the underlying discriminative structure of the features. This trained denoiser serves as the basic component of the proposed diffusion-based defense, providing purified feature representations that are later used for robust classification.
{
The identical figures in this subsection visually demonstrate the denoising behavior, illustration that the denoiser output shifts closer to the original clean feature representation and that the reconstruction error is reduced compared to the noisy input.
}
\begin{figure}
    \centering
    \includegraphics[width=0.85\textwidth]{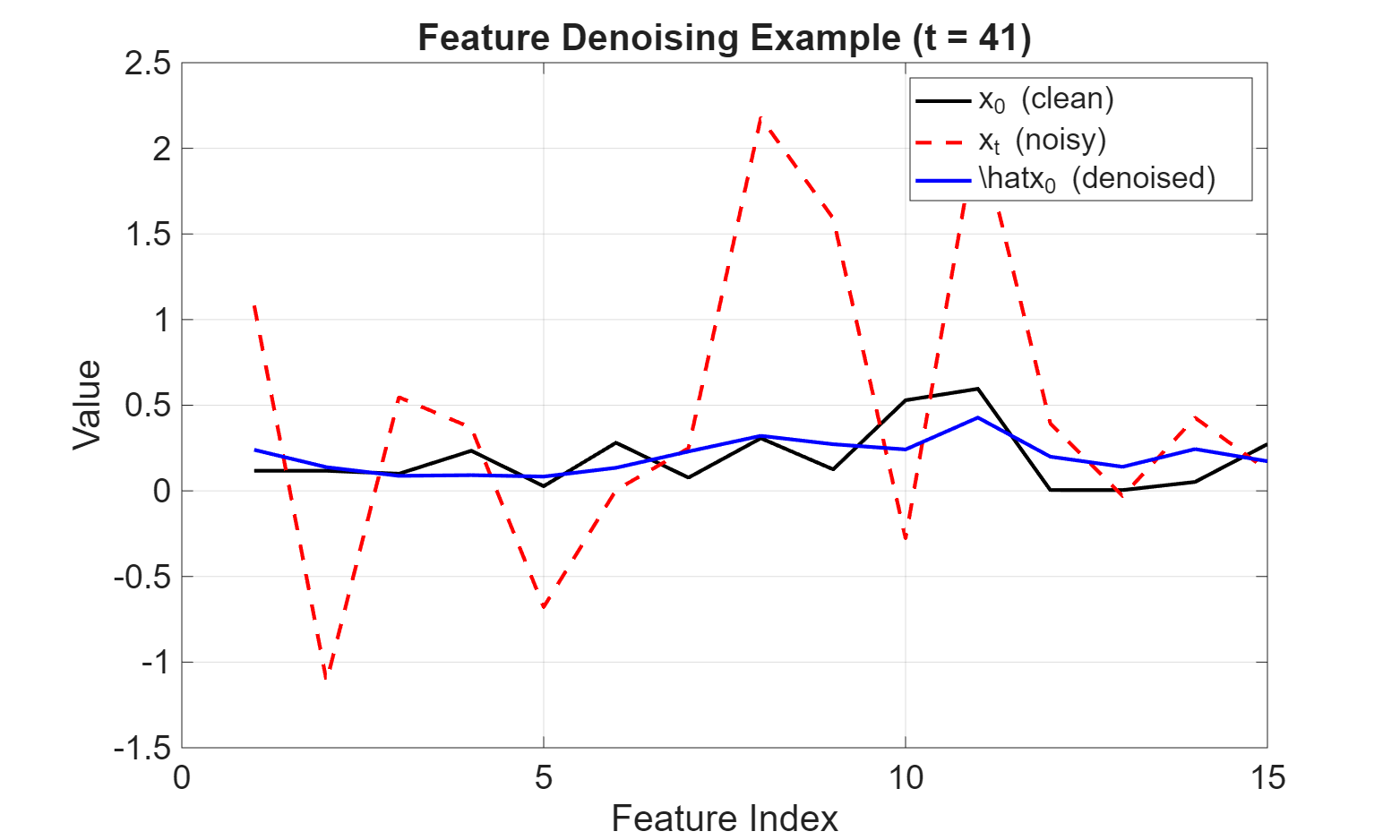}
    \caption{Feature denoising example at diffusion step t=41. The clean feature vector X0, its diffused version xt, and the denoiser output x0\^{} are drawn to explain how the network suppresses diffusion noise and movement, acting closer to the clean features.}
    \label{fig:nnmf_feature06}
\end{figure}
\begin{figure}
    \centering
    \includegraphics[width=0.85\textwidth]{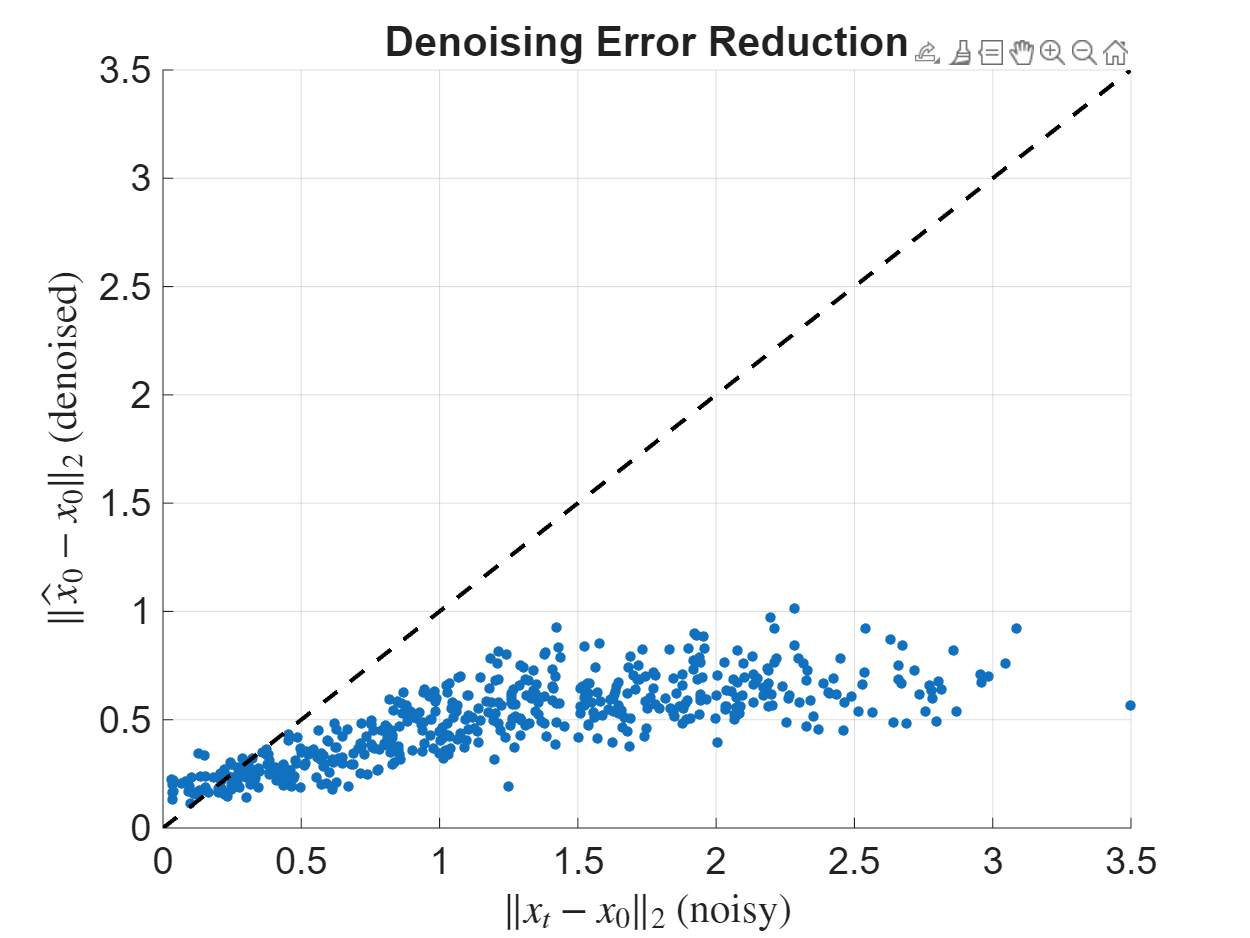}
    \caption{Denoising error is lowering relative to the noisy information. Each point contrast the noisy reconstruction error, $\|x_t - x_0\|_2$ (x-axis), versus the denoised reconstruction error, $\|\hat{x}_0 - x_0\|_2$ (y-axis). Points lying below the identity streak indicate successful fault decrees after denoising, demonstrating that the denoiser effectively returns features near the original clean representation.}
    \label{fig:nnmf_feature06}
\end{figure}
\begin{figure}
    \centering
    \includegraphics[width=0.85\textwidth]{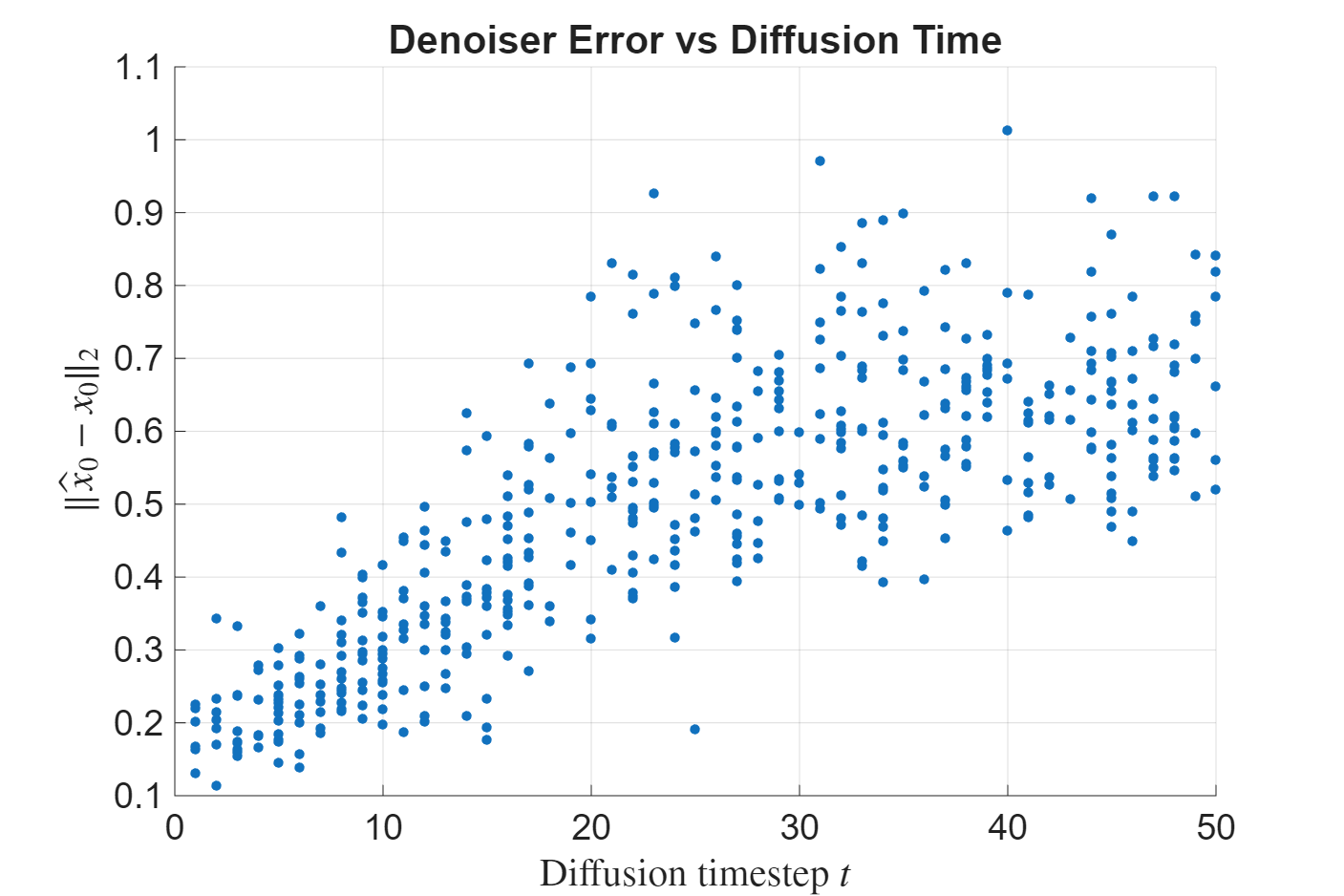}
    \caption{Denoiser reconstruction error versus diffusion time. The plot appears $x_0\|2\| \hat{x}_0$ as a function of timestep $t$, highlighting how denoising hard changes with increasing diffusion force.}
    \label{fig:nnmf_feature06}
\end{figure}

\subsection{Evaluation of Feature-Space Diffusion Refinement}
In this stage, we evaluate the suggested feature-space diffusion defense at the test time. The method implemented a forward diffusion step to perturb the clean NNMF feature vectors at a chosen timestep purt, then used the trained denoiser to assess the clean features x0. Finally, we compare the classifier performance on clean vs purified (defended) features using confusion matrices and test accuracy, and export a Python-ready defended test bundle for AutoAttack.
 The following data tables report types and quantitative comparisons between clean and diffusion-defended test features.
 {
The figures in this subsection appear clean and defended feature representations, confusion matrices, and standard test accuracy, allowing a direct visual assessment of the effect of diffusion-based refinement on classification results.
}
\begin{figure}
    \centering
    \includegraphics[width=0.85\textwidth]{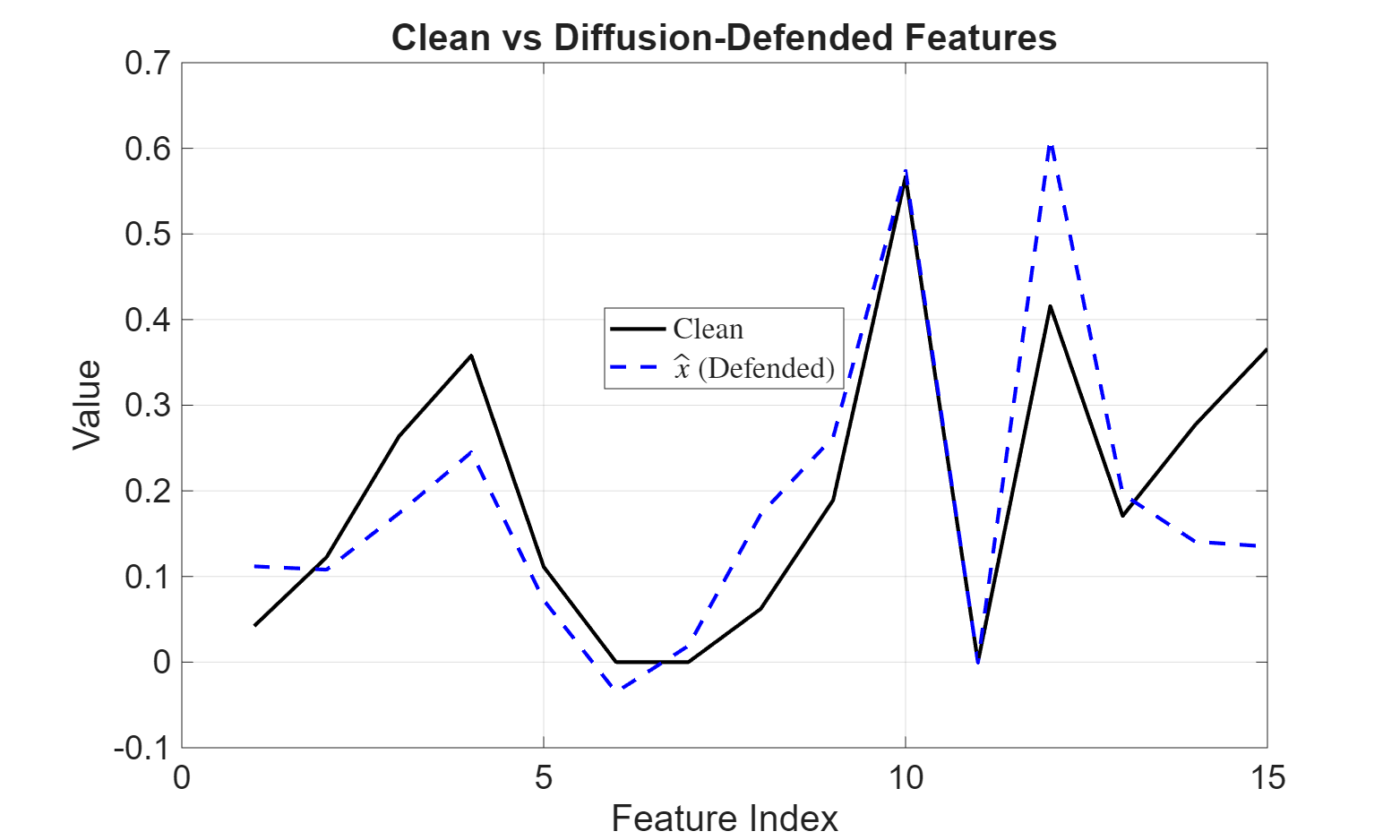}
    \caption{Clean vs. diffusion-defended NNMF feature vector example on the test set, explain how the refine step alters the feature profile after forward noise and denoising at purt .}
    \label{fig:nnmf_feature06}
\end{figure}
\begin{figure}
    \centering
    \includegraphics[width=0.85\textwidth]{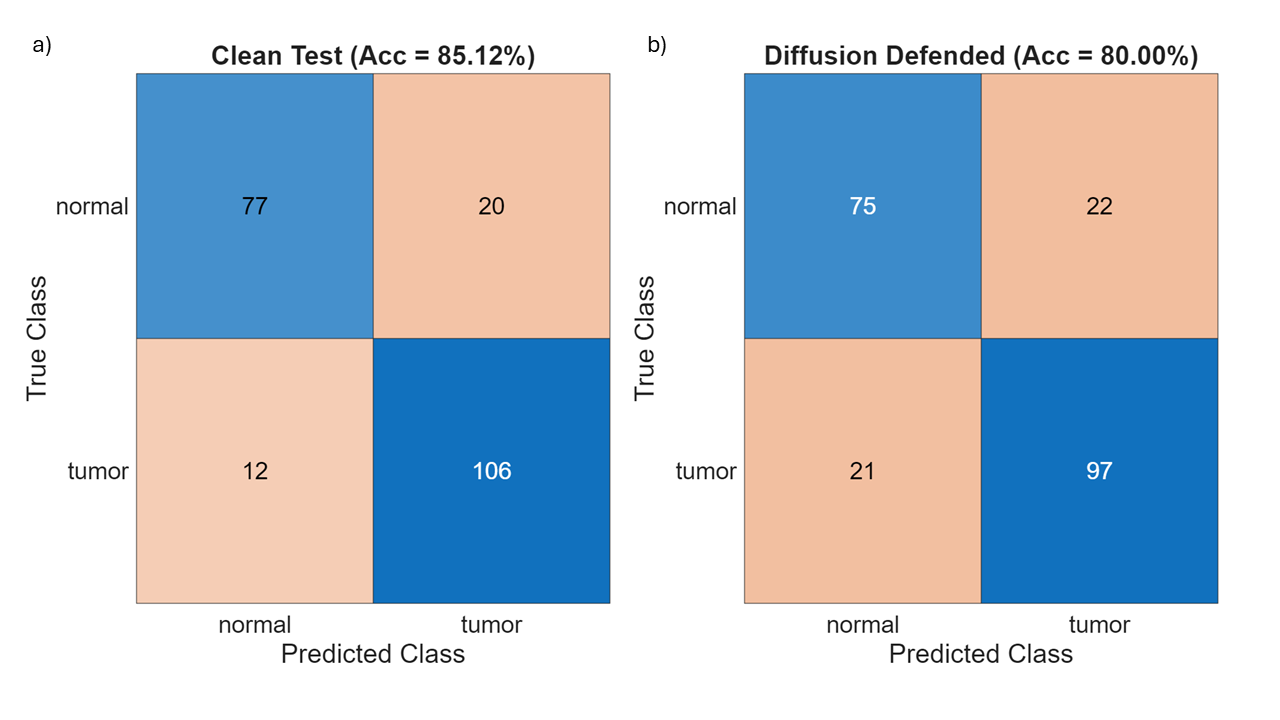}
    \caption{a) Confusion matrix on the test set using clean (non-defended) NNMF features, displays class-wise predicate outcomes for normal and tumor. b) Confusion matrix on the test set after implementation of diffusion-based feature refinement (defended features), highlighting changes in misclassification patterns compared to the clean state.}
    \label{fig:nnmf_feature06}
\end{figure}
\begin{figure}
    \centering
    \includegraphics[width=0.85\textwidth]{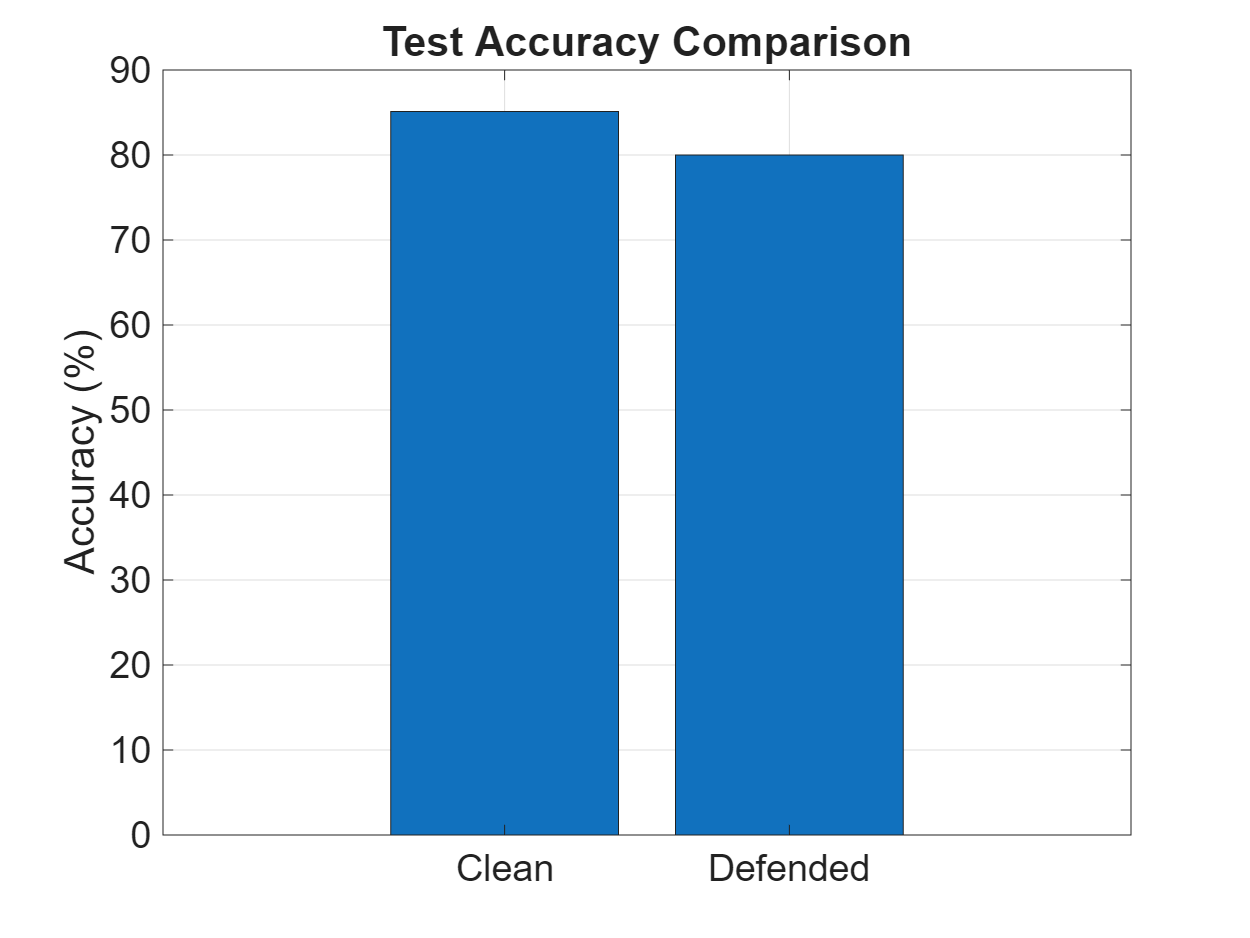}
    \caption{Test accuracy comparison between clean features and diffusion-defended (refine) features, quantifying the net effect of the defense on standard classification accuracy.}
    \label{fig:nnmf_feature06}
\end{figure}
\subsection{Robustness Evaluation under AutoAttack}
In this stage, we evaluate the robustness of the suggested diffusion-based feature defense under robust adversarial attacks. We run AutoAttack ($L\infty$, $\epsilon=0.10$ ) using two components (APGD-CE and Square) on both the clean classifier and the defended pipeline (forward diffusion noise + feature denoiser + classifier). Since defense is random due to injected noise, Expectation over Transformation (EOT) is applied by medium predictions over multiple random samples (K=8). The robustness is according to the attack and as the final robust accuracy.
{
The figures in this part summarize the robustness evaluation under AutoAttack, including clean-versus-robust accuracy, robust accuracy per-attack, and the reduction in performance drop achieved by the defended pipeline compared to the baseline model. The baseline refered to the same NNMF\&CNN pipeline without the diffusion defense, thus the baseline was the same architecture but undefended.
}
\begin{figure}
    \centering
    \includegraphics[width=0.85\textwidth]{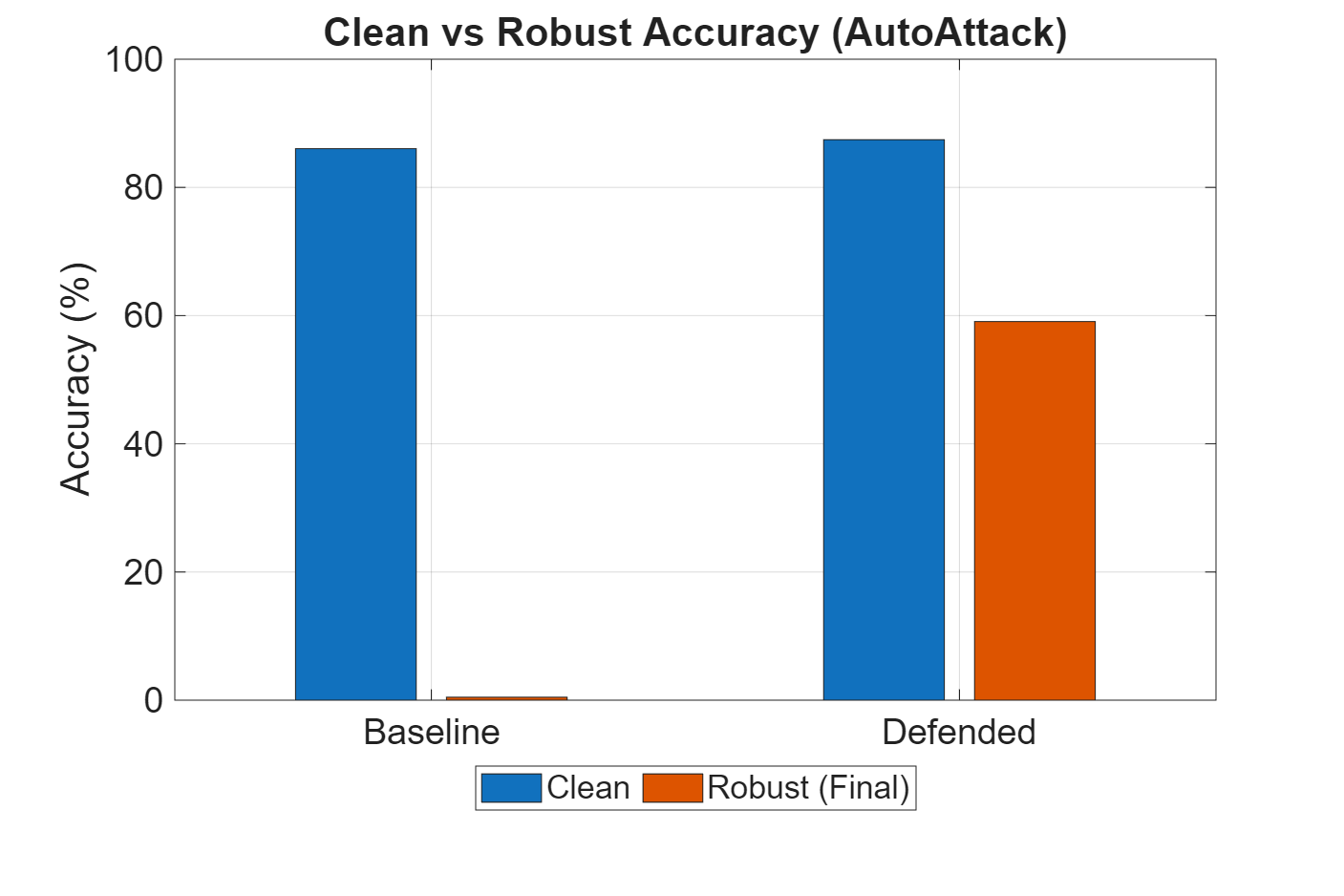}
    \caption{Clean and robust accuracy under AutoAttack ($L\infty$, $\epsilon=0.10$) for the clean model and the suggested diffusion-based defense. The robust accuracy corresponds to the final AutoAttack score (minimum across the evaluated attacks).}
    \label{fig:nnmf_feature06}
\end{figure}
\begin{figure}
    \centering
    \includegraphics[width=0.85\textwidth]{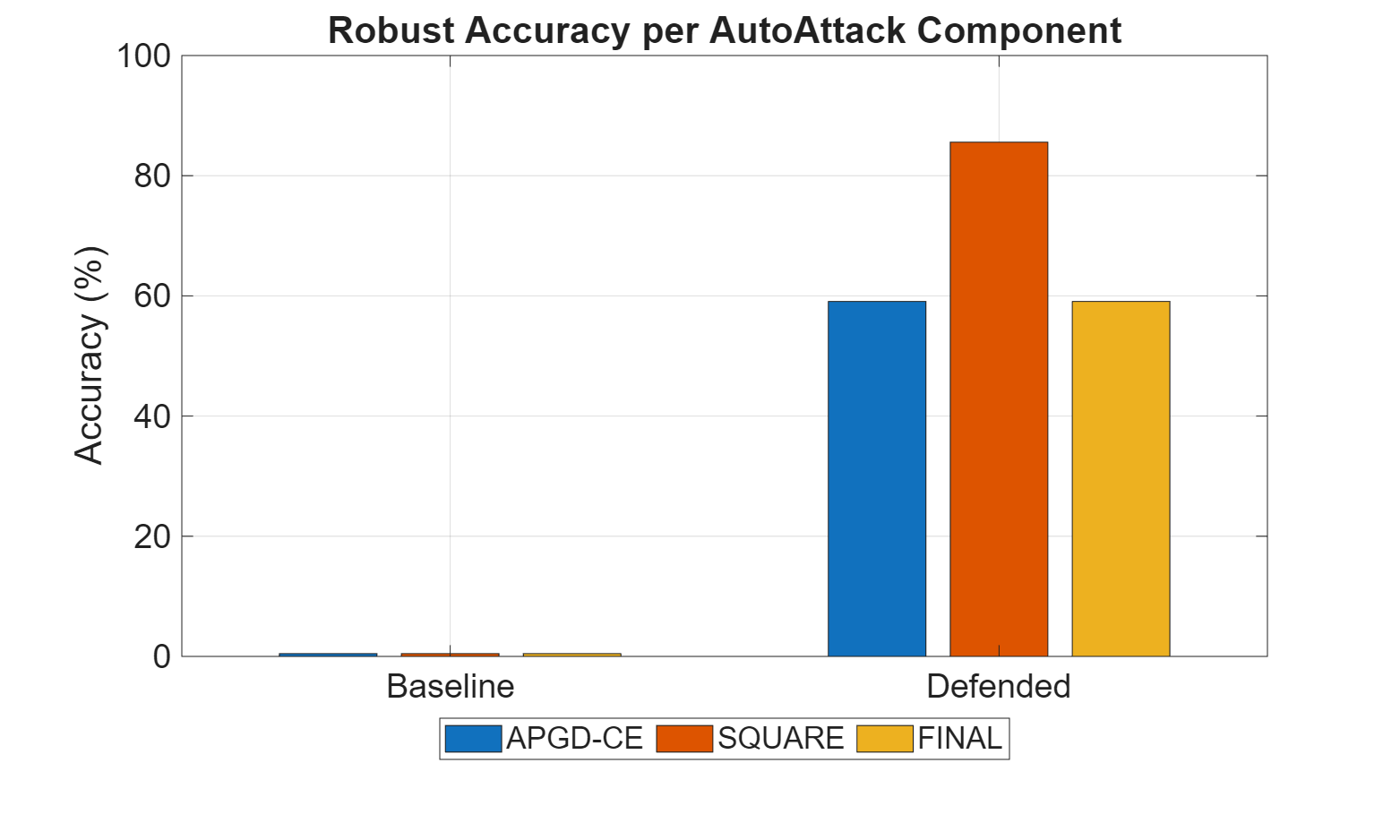}
    \caption{Robust accuracy per AutoAttack component (APGD-CE and Square) for the baseline and defended models (L$\infty$, $\epsilon=0.10$). The final robustness is calculated as the minimum accuracy across attacks.}
    \label{fig:nnmf_feature06}
\end{figure}
\begin{figure}
    \centering
    \includegraphics[width=0.85\textwidth]{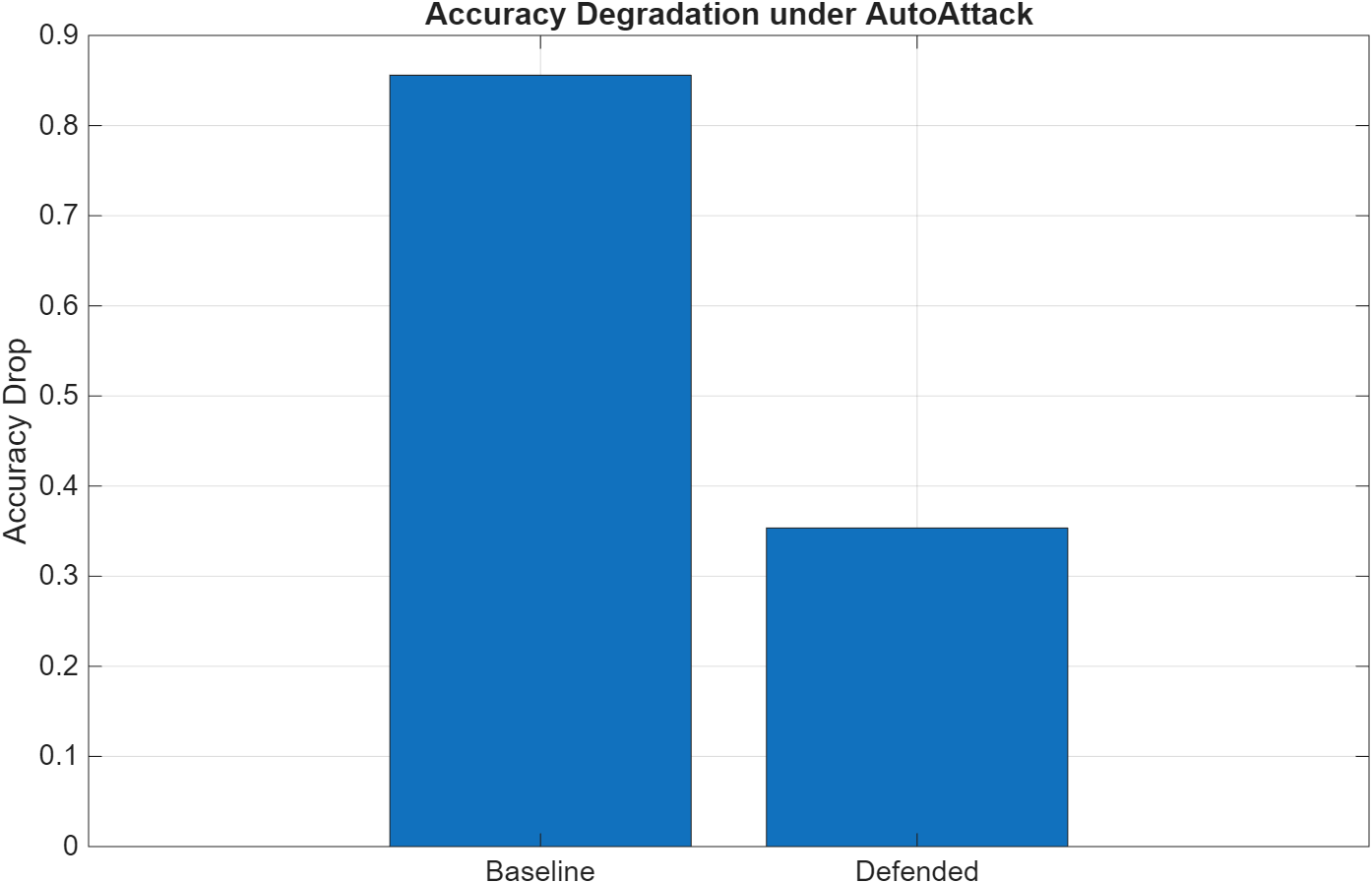}
    \caption{Accuracy decline under AutoAttack for baseline versus defended models (L$\infty$, $\epsilon$=0.10). The diffusion-based defense noticeably reduces the drop from clean to robust performance.}
    \label{fig:nnmf_feature06}
\end{figure}
\subsection{Comprehensive Performance Evaluation}

In addition to classical classification metrics, a total probabilistic 
rating was performed to estimate discrimination capacity and probability 
standardization under clean and adversarial conditions. Inspired by practical  discussions on metric selection in the classification system~\cite{medium2023metric}, 
We used ROC-AUC, Brier score and Log-Loss, along with precision, 
Recall, F1-score, Matthews Correlation Coefficient (MCC) 
and balanced precision.

\subsubsection{ROC-AUC}

The Receiver Operating Characteristic Area Under Curve Measurement (ROC-AUC)
 measures how well the pattern states positive samples above negative samples, 
independently of a fixed classification threshold ~\cite{medium2023metric}. 
A higher ROC-AUC indicates a powerful discrimination ability.

\subsubsection{Brier Score}

The Brier score estimates the average squared error of the predicted probabilities:

\begin{equation}
\text{Brier Score} = 
\frac{1}{N} \sum_{i=1}^{N} (p_i - y_i)^2
\end{equation}

This metric can standardize the probability, where lower values indicate 
the highest probabilistic accuracy.

\subsubsection{Log-Loss}

Log-Loss (cross-entropy loss) calculates the chance of predicted probabilities 
Given the true labels:

\begin{equation}
\text{LogLoss} =
-\frac{1}{N} \sum_{i=1}^{N}
\left[
y_i \log(p_i) + (1-y_i)\log(1-p_i)
\right]
\end{equation}

Log-Loss with difficulty penalizes overconfident untrue predictions, making it 
useful and informative in adversarial tuning~\cite{medium2023metric}.
\subsubsection{Matthews Correlation Coefficient (MCC)}

The Matthews Correlation Coefficient (MCC) is a stable performance metric 
that considers all values of the confusion matrix and is particularly 
suitable for unbalanced datasets. It is defined as:

\begin{equation}
\text{MCC} =
\frac{TP \times TN - FP \times FN}
{\sqrt{(TP+FP)(TP+FN)(TN+FP)(TN+FN)}}
\end{equation}

where $TP$, $TN$, $FP$, and $FN$ denote true positives, true negatives, 
false positives, and false negatives, respectively. 
The MCC amount ranges from $-1$ (total disagreement) to $+1$ (perfect prediction)

Balanced Accuracy was also included to relieve the possibility of class imbalance effects.
\begin{equation}
\text{Balanced Accuracy} = \frac{Sensitivity + Specificity}{2}
\end{equation}

\subsubsection{Results and Discussion}
The results show that while the clean model suffers substantial degradation under adversarial attacks, the suggested diffusion-based defense preserves strong discriminative capability (ROC-AUC) and improved probability calibration (lower Brier 
Score and Log-Loss). These results support the importance of evaluating 
the classification model that employs discrimination and probabilistic metrics, 
as highlighted in~\cite{medium2023metric}. The approximation −1 of MCC under AutoAttack perturbations for the baseline indicates a complete reveal of the predictions and confirms that the adversarial attack effectively destroyed the classifier.
Table~\ref{tab:comprehensive_metrics} summarizes the quantitative comparison.
{
It is important to explain that the baseline model is designed purposely as a standard reference without any defense technique. Its role is not to achieve maximum robustness, but to provide a fair comparison framework to evaluate the effectiveness of the proposed diffusion-based defense.

The performance degradation observed in the baseline model under adversarial attacks highlights its vulnerability, which is consistent with the widely reported behavior of deep learning models in the literature.

In contrast, the proposed defended model demonstrates significantly improved robustness under the same attack conditions, confirming that the performance gain is directly attributed to the introduced diffusion-based feature purification rather than architectural complexity.

This comparison ensures a fair and controlled evaluation in which the contribution of the proposed defense can be clearly isolated and validated.

Although the baseline model displays a considerable performance drop under AutoAttack (Acc = 0.0047), this attitude is regular with a widely reported sensitivity of deep neural networks under powerful adversarial attacks. The objective of the baseline is to provide a standard reference without any defense technique.

Importantly, the improvement achieved by the proposed method is reflected not only in relative gains, but also in absolute robustness performance (Acc = 0.5953), which demonstrates substantial resilience under the same attack conditions. This confirms that the observed improvement is not exaggerated, but reflects the effectiveness of the proposed diffusion-based defense.
}

\begin{table}
\centering
\caption{Comprehensive Performance Comparison Under Clean and Adversarial Settings}
\label{tab:comprehensive_metrics}
\begin{tabular}{lccccccccc}
\toprule
Model & Acc & Prec & Rec & F1 & MCC & BalAcc & ROC-AUC & Brier & LogLoss \\
\midrule

Clean\_Baseline 
& 0.8605 & 0.8548 & 0.8983 & 0.8760 & 0.7178 & 0.8564 & 0.9105 & 0.1461 & 0.4751 \\

Clean\_Defended 
& 0.8512 & 0.8525 & 0.8814 & 0.8667 & 0.6988 & 0.8479 & 0.8967 & 0.1555 & 0.4963 \\

Robust\_Baseline 
& 0.0047 & 0.0000 & 0.0000 & 0.0000 & -0.9906 & 0.0052 & 0.0075 & 0.4702 & 1.1629 \\

Robust\_Defended 
& 0.5953 & 0.6115 & 0.7203 & 0.6615 & 0.1703 & 0.5818 & 0.7485 & 0.2150 & 0.6182 \\

\bottomrule
\end{tabular}
\end{table}
{
To improve readability, key results in clean, adversarial and defended scenarios are summarized in a unified comparison format, allowing for direct evaluation of performance differences.
}

\begin{figure}[!ht]
\centering
\includegraphics[width=\columnwidth]{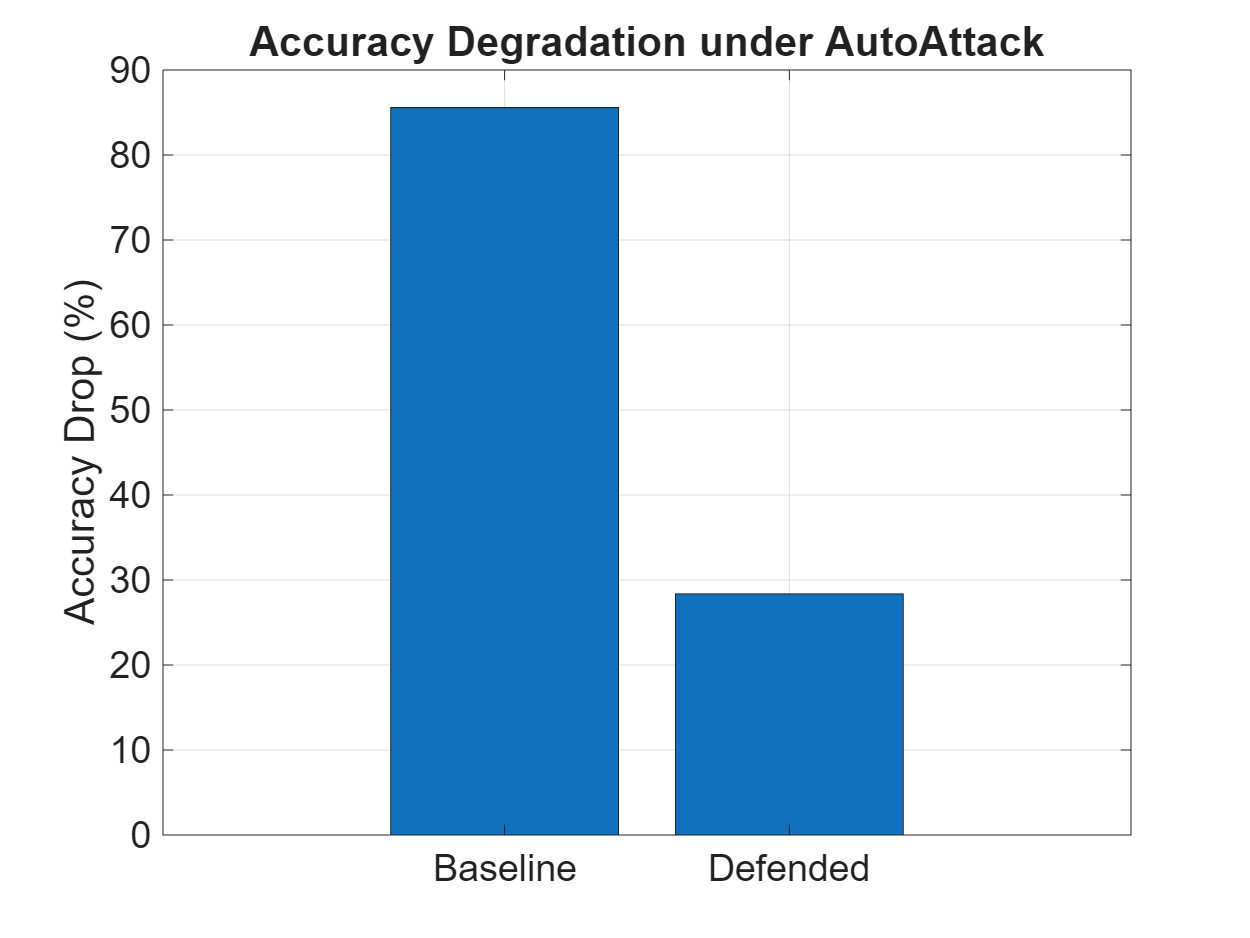}
\caption{Comparison of classification and probabilistic metrics for baseline 
    and diffusion-based defense under clean and adversarial settings. 
    Higher values indicate better performance except for Brier Score and Log-Loss, 
    where lower values are preferred.}
\label{fig:metrics_comparison}
\end{figure}

Figure~\ref{fig:metrics_comparison} illustrates the comparative behavior of 
the evaluated metrics.
{
The figure provides an integrated comparison of standard classification and probabilistic evaluation metrics in clean and adversarial settings, confirming that the proposed diffusion-based defense preserves a stronger overall performance than the undefended baseline under attack.
}


{
A comparison with recent studies shows that, while some procedures have cleaner accuracy, they often lack robustness and interpretability. The suggested method provides a balanced trade-off between accuracy, robustness, and interpretability, which distinguishes it from existing models.
}

\section{Implementation and Computational Performance Analysis}

The suggested framework was running using a combined MATLAB--Python pipeline. 
MATLAB was applied to NNMF for the extraction of features, statistical classification, 
CNN training, and a diffusion-based purification system. 
The trained models were exported to the ONNX format and run in Python 
using PyTorch and the AutoAttack framework for adversarial robustness rating.

All tests were carried out on a workstation armed with an 
Intel Core i7 CPU (16 GB RAM) and an NVIDIA RTX-series GPU 
with CUDA acceleration.

\subsection{Computational Time Comparison}

To analyze computational efficiency, we measured the execution time for all main phases of the suggested framework in both CPU 
and GPU environments. The results are summarized in 
Table~\ref{tab:execution_time_comparison}.

\begin{table}
\centering
\caption{Execution Time Comparison (CPU vs GPU)}
\label{tab:execution_time_comparison}
\begin{tabular}{lcc}
\hline
\textbf{Stage} & \textbf{CPU (sec)} & \textbf{GPU (sec)} \\
\hline
NNMF Feature Extraction & 17.03 & 9.11 \\
CNN Training (NNMF Features) & 12.60 & 17.22 \\
Diffusion Denoiser Training & 8.42 & 9.24 \\
AutoAttack Baseline (APGD-CE + Square) & 8.42 & 10.73 \\
AutoAttack Defended (APGD-CE + Square) & 155.00 & 70.30 \\
\hline
\textbf{Total Runtime} & \textbf{201.47} & \textbf{116.60} \\
\hline
\end{tabular}
\end{table}

\subsection*{Analysis}

The results explain that GPU acceleration basically reduces 
the total runtime of the suggested framework. In special, 
the adversarial robustness estimate phase shows the most 
significant improvement, where the defended AutoAttack runtime 
minimizes from 155.00 seconds on CPU to 70.30 seconds on GPU, 
and there appears to be a major computational load.

The main differences are observed in lightweight stages, such as 
CNN training, due to GPU initialization and kernel beginning overhead. 
Since the NNMF feature volume is relatively small, the GPU does 
not provide a consistent advantage for all phases.

In general, GPU acceleration improves the overall runtime from 
201.47 seconds to 116.60 seconds, obtaining approximately 
a speedup of 1.73× speedup. These results show that hardware acceleration 
is particularly useful for computationally strong adversarial 
robustness estimates, making the proposed framework practical feasible for large-scale analysis.

\section{Conclusion}
This study approaches a robust and structured framework for brain tumor classification based on NNMF feature extraction, statistical feature
selection, CNN-based classification, and diffusion-based feature
purification. Unlike a traditional end-to-end deep learning example.
that builds just on high-dimensional image input, the proposed
path uses interpretable low-rank NNMF representations to
build a compact and discriminative feature space.

The extracted NNMF components were statistically rated using
metrics such as AUC, Cohen’s $d$ and hypothesis testing, allowing
The choice of the most informative features. A lightweight CNN
classifier trained on the selected feature subset demonstrated
competitive execution in clean test data while ensuring 
computational efficiency.

To direct adversarial vulnerability, a feature-space diffusion
technique  was applied, followed by a learned denoising model
that approximates the reverse diffusion process. The
robustness estimate under powerful adversarial attacks using AutoAttack
guaranties that the proposed defense enhances stability compared
to the baseline classifier. The experimental results show that
combining interpretable NNMF features with diffusion-based
purification improves adversarial robustness while maintaining
classification accuracy.

In general, the suggested framework explains that the
combination of statistical feature analysis, compact neural architectures,
and a structured feature-level defense technique provides a
stable solution between interpretability, accuracy, and
robustness in the medical image classification model.

Future work may check the adaptive diffusion table,
multi-step purification strategies, and extension to
multi-class tumor levels tasks.

\section{Acknowledgement}
This work was supported by the Distinguished Professor Program of Óbuda University. The authors are also grateful for the possibility of using the HUN-REN Cloud \url{https://science-cloud.hu/en}~\cite{H_der_2022} which helped us achieve some particular results published in this paper. Hiba Adil Al-kharsan gratefully acknowledges the financial support of the Stipendium Hungaricum Doctoral Program, managed by the Tempus Public Foundation.
{
	During the preparation of this manuscript, the authors used artificial intelligence to improve the language and clarify the explanations, the authors checked and edited all the results and take full responsibility for the content of this manuscript.}

\bibliography{references}

\end{document}